\documentclass[journal]{IEEEtran}
\usepackage[T1]{fontenc}
\ifCLASSINFOpdf
\else
\fi

\usepackage{cuted}
\usepackage{graphicx}
\usepackage{amsfonts}
\usepackage{amsmath}
\usepackage[colorlinks,linkcolor=red,anchorcolor=blue,citecolor=green]{hyperref}
\usepackage{listings}
\usepackage{amssymb}
\usepackage{amsthm}
\usepackage{graphicx}
\usepackage{epstopdf}
\usepackage{setspace}
\usepackage{booktabs}
\usepackage{threeparttable}
\usepackage{multirow}
\usepackage[mathlines]{lineno}
\usepackage{subfigure}
\usepackage[linesnumbered,ruled,vlined,algo2e]{algorithm2e}
\usepackage{threeparttable}
\usepackage{mathrsfs}
\theoremstyle{plain}

\theoremstyle{definition}
  \newtheorem{defn}{Definition}
  
  \newtheorem{exmp}{Example}
  \newtheorem{rem}{Remark}

\theoremstyle{break}

\theoremstyle{definition}

\begin{document}
%
\title{\huge Generation of Granular-Balls for Clustering 
Based on the Principle of Justifiable Granularity}
%
%
%

\author{Zihang~Jia, Zhen~Zhang, \IEEEmembership{Senior Member, IEEE}
and Witold Pedrycz, \IEEEmembership{Life Fellow, IEEE}
\thanks{This work was partly supported by the National Natural Science Foundation of China under Grant Nos. 72371049 and 71971039, the Funds for Humanities and Social Sciences of Ministry of Education of China under Grant No. 23YJC630219 and the Fundamental Research Funds for the Central Universities of China under Grant No. DUT23RW406.
(\emph{Corresponding author: Zhen Zhang.})}
\thanks{Zihang Jia and Zhen Zhang are with the Institute of Systems Engineering, School of Economics and Management, 
Dalian University of Technology, Dalian 116024, China (e-mails:
zihangjia@outlook.com; zhen.zhang@dlut.edu.cn).
}
\thanks{Witold Pedrycz is with the Department of Electrical and Computer Engineering,
University of Alberta, Edmonton, AB T6R 2V4, Canada, with the Systems 
Research Institute, Polish Academy of Sciences, Warsaw, Poland, and Istinye 
University, Faculty of Engineering and Natural Sciences, Department of 
Computer Engineering, Sariyer/Istanbul, Turkiye (e-mail: 
wpedrycz@ualberta.ca). 
}
}

\maketitle
\begin{abstract}

Efficient and robust data clustering remains a challenging task in the field of data analysis. Recent efforts have explored the integration of granular-ball (GB) computing with clustering algorithms to address this challenge, yielding promising results. However, existing methods for generating GBs often rely on single indicators to measure GB quality and employ threshold-based or greedy strategies, potentially leading to GBs that do not accurately capture the underlying data distribution. To address these limitations, this article introduces a novel GB generation method. The originality of this method lies in leveraging the principle of justifiable granularity to measure the quality of a GB for clustering tasks. To be precise, we define the coverage and specificity of a GB and introduce a comprehensive measure for assessing GB quality. Utilizing this quality measure, the method incorporates a binary tree pruning-based strategy and an anomaly detection method to determine the best combination of sub-GBs for each GB and identify abnormal GBs, respectively. Compared to previous GB generation methods, the new method maximizes the overall quality of generated GBs while ensuring alignment with the data distribution, thereby enhancing the rationality of the generated GBs. Experimental results obtained from both synthetic and publicly available datasets underscore the effectiveness of the proposed GB generation method, showcasing improvements in clustering accuracy and normalized mutual information.
\end{abstract}


%
\IEEEpeerreviewmaketitle

\section{Introduction}\label{Section1}
\IEEEPARstart{C}{lustering} analysis, which aims to assign unlabeled instances into clusters so that instances within the same cluster are more similar to each other than to those in different clusters, is a fundamental unsupervised learning task in machine learning \cite{Jain1999}. It has been widely used in various fields such as image segmentation \cite{Lei20191753, Hou20163182}, time series analysis \cite{Guo2021, Guo2024903, Yang20237622, Guo2022755}, and text mining \cite{Guan2022}. Over the past decades, several clustering algorithms have been developed, including $k$-means clustering \cite{Macqueen1967}, $k$-means++ clustering \cite{Arthur20071027}, spectral clustering (SC) \cite{VonLuxburg2007395}, and density peaks clustering (DPC) \cite{Rodriguez2014}. While these algorithms are effective, they may not efficiently handle increasingly large-scale data robustly. For example, the $k$-means clustering algorithm, though simple and fast, struggles to identify non-spherical clusters and often converges to local optima \cite{Omran2007583}. While SC algorithms offer a solution by transforming the clustering problem into a graph partitioning problem, the high time complexity of constructing affinity matrices and solving the eigen-decomposition problem poses challenges. Consequently, how to efficiently and robustly cluster data, especially large-scale data, remains a challenging issue.

\begin{figure}[htbp]
  \centering
  \subfigure[]{\includegraphics[width=1.5in]{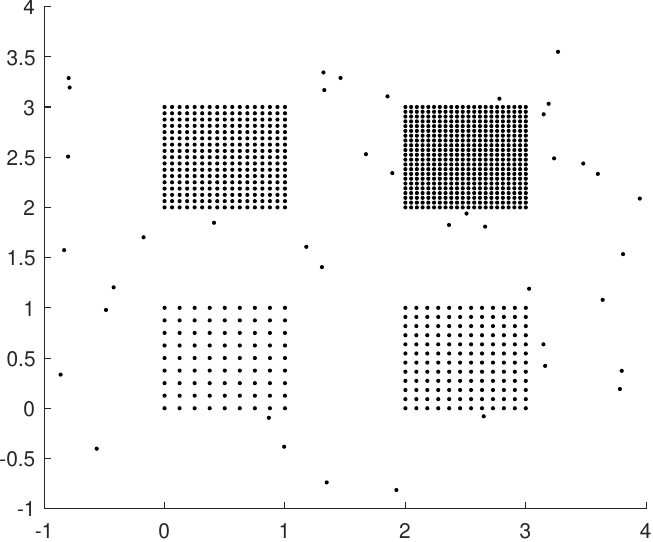}}
  \subfigure[]{\includegraphics[width=1.5in]{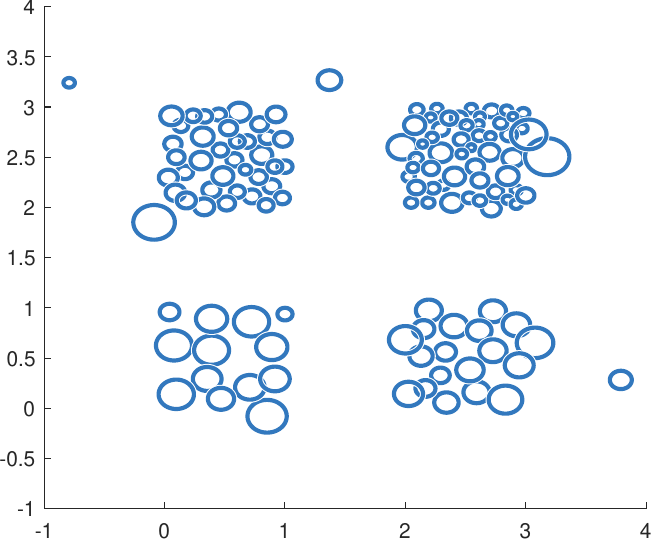}}
  \caption{Representing instances through GBs. (a) A synthetic dataset; 
  (b) The GBs used to represent the synthetic dataset.
  }
  \label{Fig_Example_GB}
\end{figure}

Recently, granular-ball (GB) computing has been integrated with existing clustering algorithms to address the challenges encountered in clustering methods, achieving promising performance \cite{ChengDongDong2023,XieJiang2023,CHENG2024123313}.
In essence, GB computing is a novel granular computing approach characterized by its simple formulation, initially proposed by Xia et al. in \cite{XIA2019136}.
Grounded on the fundamental assumption that any instance and its local neighbors are highly likely to belong to the same cluster \cite{ChengDongDong2023}, GB computing treats similar instances as a collective entity represented by a distinct information granule termed GB, which manifests as a hypersphere defined by a center and a radius (see, for instance, Fig. \ref{Fig_Example_GB}). Following the paradigm of GB computing, Cheng et al. \cite{ChengDongDong2023} introduced a granular-ball-based density peaks clustering (GBDPC) algorithm to enhance the DPC algorithm. GBDPC only requires calculating the distance between GBs and the density of each GB, rather than the distance between individual instances and the density of each instance. Compared to the DPC algorithm, GBDPC is not only more effective but also parameter-free. Similarly, to improve the efficiency of SC algorithms, Xie et al. \cite{XieJiang2023} proposed the granular-ball-based spectral clustering (GBSC) algorithm, which substitutes the similarity matrix of instances with the similarity matrix of GBs. In \cite{CHENG2024123313}, the GB computing paradigm was introduced to manifold learning for clustering high-dimensional large-scale data. These algorithms offer a novel perspective on clustering within the framework of GB computing. Generally, GB-based clustering algorithms demonstrate several advantages. On the one hand, they significantly reduce computational complexity since the number of GBs is much smaller than the number of instances. On the other hand, given that the GB serves as a coarse-grained representation of instances \cite{ChengDongDong2023}, these algorithms yield more robust results.

To ensure the reliability of GB computing-based algorithms, it is crucial to select an appropriate method for generating GBs. If the generated GBs do not align with the data distribution, the calculation results of GB computing-based algorithms may become unreliable. In supervised learning, where instance labels are available, the generation of GBs relies on these labels.  Typically, the quality of a GB is assessed by its purity, which represents the percentage of instance labels that are most frequently present within the GB \cite{XIA2019136}. To ensure that the generated GBs align with the data distribution, it is essential to guarantee that the purity of any GB exceeds a certain threshold, such as 0.99.  Given the intuitive and effective nature of using purity to evaluate the quality of a GB, GB computing has found successful application in various supervised learning tasks, including classification \cite{XIA2019136,Peng2022504}, incremental learning \cite{Li2023,ZhangQinghua2023}, sampling \cite{XiaShuyin2023}, and feature selection \cite{XiaShuyin2022,Qian2023,QIAN2023119698}.
However, in the context of clustering, it may be challenging to generate appropriate GBs due to the absence of instance labels within the dataset.  In the studies conducted by Cheng et al. \cite{ChengDongDong2023,CHENG2024123313}, it was postulated that if the number of instances belonging to a particular GB is sufficiently small, then the majority of these instances are likely to belong to the same cluster. Based on this assumption, the quality of a GB is evaluated primarily by the number of instances it contains. Specifically, a GB is considered qualified if the count of its constituent instances falls below a predefined threshold. In contrast, Xie et al. \cite{XieJiang2023} employed a different metric called the distribution measure to assess the quality of a GB, which is essentially the average distance between each instance and the center of the GB. To optimize the overall quality of the generated GBs, a greedy strategy is adopted in \cite{XieJiang2023}. Despite the successful application of these GB generation methods, two primary challenges remain:

\begin{enumerate}
  \item [(1)] For clustering tasks, the absence of instance labels makes it challenging to comprehensively and effectively measure the quality of a GB using a single indicator. On one hand, some GBs generated by the metrics defined in \cite{ChengDongDong2023,CHENG2024123313} may be deemed unqualified due to their inclusion of a large number of instances. On the other hand, relying solely on distribution measures to assess GB quality might result in favoring GBs with only a single instance, which contradicts the fundamental assumption of GB computing that aims to represent instances using GBs. Therefore, there is a critical need to introduce new metrics that evaluate GB quality from multiple perspectives.
  \item [(2)]   Previous quality metrics utilize thresholds \cite{ChengDongDong2023,CHENG2024123313} or a greedy strategy \cite{XieJiang2023} to generate GBs in clustering tasks. However, these approaches do not guarantee the overall quality of the generated GBs, and there is a risk that the generated GBs may not accurately align with the distribution of the data.  To address this challenge, it is imperative to devise novel strategies aimed at maximizing the quality of generated GBs, thereby ensuring their consistency with the underlying data distribution.

\end{enumerate}

The main aim of this article is to present a new GB generation method for clustering tasks,  aimed at tackling the aforementioned challenges. The key focus in achieving this objective lies in devising a robust approach for measuring the quality of a GB. Fortunately, the principle of justifiable granularity (POJG) \cite{PEDRYCZ20134209} offers a comprehensive framework for designing information granules, even in scenarios where instance labels are unavailable. According to POJG, an information granule should not only accurately represent the underlying data but also possess well-defined semantics. Thus, in this article, we innovatively leverage the POJG to assess the quality of GBs specifically tailored for clustering tasks. In this way, 
the generation process of GBs can be enhanced such that generated GBs are more suitable for clustering tasks.
The key contributions of this article are outlined as follows:
\begin{enumerate}
  \item [(1)] Following the POJG, we propose a novel method to comprehensively measure the quality of a GB for clustering tasks, taking into account both specificity and coverage. This approach allows for a more comprehensive assessment of GB quality.
      
  \item [(2)] To maximize the overall quality of generated GBs, we introduce the best quality measure and the best combination of sub-GBs for each GB, along with a corresponding calculation method. This approach maximizes the overall quality of the generated GBs, resulting in representations that are not only of higher quality compared to existing methods but also capture data at a coarser granularity.
  \item [(3)] To enhance the rationality of the generated GBs and mitigate the impact of abnormal GBs, we redefine abnormal GBs based on the POJG and conduct anomaly detection for the generated GBs.
  \item [(4)]  By integrating these procedures, we propose a novel and effective GB generation method based on the POJG for clustering tasks, effectively addressing existing challenges in GB-based clustering algorithms.
\end{enumerate}

The remainder of this article is structured as follows: In Section \ref{Section2}, we provide a comprehensive and focused review of related works. Section \ref{Section3} presents the proposed GB generation method for clustering tasks based on the POJG. Following this, Section \ref{Section4} covers a series of numerical experiments conducted on both synthetic and publicly available datasets to validate the performance of the proposed GB generation method. Finally, Section \ref{Section5} offers concluding remarks and identifies future directions.

\section{Related Works}\label{Section2}
In this section, we review some related works concerning GB computing and the POJG.

\subsection{Granular-Ball Computing for Clustering}\label{Section2.1}
To begin with, we review some basic definitions in GB computing as follows.

\begin{defn}(See \cite{XIA2019136}.)
Let $X=\{x_{1},x_{2},\cdots,x_{n}\}\subset\mathbb{R}^{m}$ be a collection of instances. The GB derived from $X$ with the center $C_{X}$
and the radius $R^{\text{Ave}}_{X}$ is denoted as $\Omega_{X},$ defined by
\begin{align}\label{Eq_Center_Average_Radius}
    C_{X}=\frac{1}{n}\sum_{i=1}^{n}x_{i};~
  R^{\text{Ave}}_{X}=\frac{1}{n}\sum_{i=1}^{n}\Vert x_{i}-C_{X}\Vert,
\end{align} 
where $\Vert\cdot\Vert$ is the $2$-norm. For any $x\in X,$ the instance $x$ is always said to belong to the GB $\Omega_{X}.$
\end{defn}

In \cite{ChengDongDong2023,XieJiang2023,CHENG2024123313}, the radius of a GB $\Omega_{X}$ is defined as the maximum distance between the
instance $x\in X$ and the center $C_{X}$ rather than the average distance $R_{X}^{\text{Ave}}$, i.e., 
$$R^{\text{Max}}_{X}=\max_{x\in X}\Vert x-C_{X}\Vert.$$
In the following discussion, to distinguish between them,
$R_{X}^{\text{Ave}}$ and $R_{X}^{\text{Max}}$ are always referred to as 
the average radius and maximum radius, respectively, of the GB $\Omega_{X}.$

We now recall two GB generation methods for clustering, 
proposed by Cheng et al. \cite{ChengDongDong2023} and Xie et al. \cite{XieJiang2023}, 
respectively.

\subsubsection{Cheng et al.'s GB Generation Method} 
In \cite{ChengDongDong2023}, Cheng et al. proposed a method for generating GBs by iteratively dividing them using the $2$-means clustering algorithm, ensuring that each resulting GB contains a sufficiently small number of instances. Specifically, given a dataset $U \subset \mathbb{R}^{m}$, let $X$ be a subset of $U$ and $\Omega_{X}$ be a GB generated from $X$. If the cardinality of $X$, denoted as $\vert X\vert$, exceeds $\sqrt{\vert U\vert}$, then $\Omega_{X}$ is split into two GBs, $\Omega_{X_{1}}$ and $\Omega_{X_{2}}$, where $X_{1}$ and $X_{2}$ are subsets of $X$ obtained by applying the $2$-means clustering algorithm to $X$. Otherwise, $\Omega_{X}$ remains unchanged. This process is repeated starting from $\Omega_{U}$ until the number of instances belonging to each GB is less than or equal to $\sqrt{\vert U\vert}$.

\begin{rem}
In Cheng et al.'s method \cite{ChengDongDong2023},  the quality of a GB is determined by 
a single indicator, namely, the number of instances belonging to the GB.
Additionally, since the maximum number of clusters is commonly set to 
$\sqrt{\vert U\vert}$ \cite{Bezdek1998IEEETCYB}, the average number of instances belonging to each cluster is less than $\sqrt{\vert U\vert}$. Therefore, the threshold $\sqrt{\vert U\vert}$ is used to determine the generated GBs, ensuring that each GB meets the qualification criterion. 
\end{rem}

\subsubsection{Xie et al.'s GB Generation Method}
In Xie et al.'s method \cite{XieJiang2023}, the average radius of a GB serves as the distribution measure to decide whether a GB should be split into two. 
Let $U$, $X$ and $\Omega_{X}$ be defined as before. Denote
\begin{align}\label{Eq_ArgMaxXX}
  (x_{\alpha},x_{\beta})=\arg\max_{(x_{i},x_{j})\in X\times X}\Vert x_{i}-x_{j}\Vert.
\end{align}
The GB $\Omega_{X}$ can be divided into $\Omega_{X_{\alpha}}$ and 
$\Omega_{X_{\beta}},$
where 
\begin{align}\label{Eq_XAlpha_XBeta}
  X_{\alpha}=\{x_{i}\in X:\Vert x_{i}-x_{\alpha}\Vert\leq \Vert x_{i}-x_{\beta}\Vert\},~ 
  X_{\beta}=X- X_{\alpha}.
\end{align}
Hereafter, $\Omega_{X_{\alpha}}$ and $\Omega_{X_{\beta}}$ are referred to as sub-GBs of $\Omega_{X}$ in the subsequent discussions. To determine whether $\Omega_{X}$ should be divided into $\Omega_{X_{\alpha}}$ and $\Omega_{X_{\beta}}$, the weighted distribution measure of $\Omega_{X}$ was introduced in \cite{XieJiang2023} defined as  
\begin{align}\label{Eq_Weighted_Distribution_Measure}
  \overline{R}^{\text{Ave}}_{X}=\frac{\vert X_{\alpha}\vert}{\vert X\vert}R^{\text{Ave}}_{X_{\alpha}}+\frac{\vert X_{\beta}\vert}{\vert X\vert}R^{\text{Ave}}_{X_{\beta}}.
\end{align}

If $\overline{R}^{\text{Ave}}_{X}$ is less than $R^{\text{Ave}}_{X}$, then the GB $\Omega_{X}$ needs to be divided into $\Omega_{X_{\alpha}}$ and $\Omega_{X_{\beta}}$. Otherwise, the GB $\Omega_{X}$ remains unchanged. This process is repeated, starting from $\Omega_{U}$, until each GB no longer requires division. To enhance the rationality of the generated GBs and mitigate the impact of abnormal GBs, anomaly detection was incorporated into this method. Let $\Phi=\{\Omega_{X_{1}},\Omega_{X_{2}},\cdots,\Omega_{X_{t}}\}$ represent a set of GBs generated from the dataset $U$. Denote $R_{\Phi}^{\text{Ave}}$ and $R_{\Phi}^{\text{Med}}$ as the average and median, respectively, of the maximum radius of GBs in $\Phi$. Then, for each $\Omega_{Y}\in\Phi$, if it holds that $R_{Y}^{\text{Max}}>2\cdot\max\left(R_{\Phi}^{\text{Ave}},R_{\Phi}^{\text{Med}}\right)$, then $\Omega_{Y}$ is identified as an abnormal GB. Any abnormal GB needs to be divided into two new GBs using Eqs. (\ref{Eq_ArgMaxXX}) and (\ref{Eq_XAlpha_XBeta}). The process of anomaly detection is repeated until there are no abnormal GBs remaining.

Obviously, Xie et al.'s method adopts a greedy strategy to maximize the overall quality of generated GBs. Now, let's consider the inequality 
$R_{X}^{\text{Ave}}>\overline{R}_{X}^{\text{Ave}}$ used in Xie et al.'s method.  In fact, by Eqs. (\ref{Eq_Center_Average_Radius}) and 
(\ref{Eq_Weighted_Distribution_Measure}), we have
$$\overline{R}_{X}^{\text{Ave}}=\frac{1}{\vert X\vert}\left(\sum_{x\in X_{\alpha}}\Vert x-C_{X_{\alpha}}\Vert+\sum_{y\in X_{\beta}}\Vert y-C_{X_{\beta}}\Vert\right).$$
Furthermore,  by Eq. (\ref{Eq_Center_Average_Radius}), we get
\begin{align*}
  \begin{split}
    &R_{X}^{\text{Ave}}>\overline{R}_{X}^{\text{Ave}}\\
  \Leftrightarrow&
  \sum_{z\in X}\Vert z-C_{X}\Vert>\sum_{x\in X_{\alpha}}\Vert x-C_{X_{\alpha}}\Vert+\sum_{y\in X_{\beta}}\Vert y-C_{X_{\beta}}\Vert.
  \end{split}
\end{align*}
Therefore, in Xie et al.'s method, the quality of a GB and its sub-GBs is solely determined by a single indicator, namely, the total distance between each instance and the center. Similarly, whether a GB is abnormal is also determined by a single indicator related to distance.

\subsection{Principle of Justifiable Granularity}
As a novel computing paradigm for information processing, granular computing (GrC) considers information granule as a superset of some formal constructs,
such as sets, intervals, fuzzy sets \cite{ZADEH1965338}, rough sets \cite{Pawlak1982}, and shadowed sets \cite{Pedrycz1998}.  In GrC, information granules serve as the fundamental unit for computation. The POJG \cite{PEDRYCZ20134209} provides a general 
framework to construct information granules, which does not rely on any specific 
formal construct. Specifically, the construction of information granules must satisfy two essential criteria:
\begin{enumerate}
  \item [(1)] Coverage: This criterion emphasizes the accumulation of data features within the constructed information granules. For instance, considering a dataset $X=\{x_{1},x_{2},x_{3},x_{4}\}$,  if an information granule $\Omega=\{x_{1},x_{3},x_{4}\}$
 is constructed, it is deemed more justifiable than another information granule $\Omega=\{x_{1},x_{4}\}$.
  \item [(2)] Specificity: This criterion underscores the necessity for well-defined semantics within constructed information granules. For instance, if an information granule $\Omega$ represents an interval, the statement ``$\Omega=[2,4]$'' conveys more specific knowledge compared to
  ``$\Omega=[1,7]$''.
\end{enumerate}

It is evident that coverage and specificity criteria often conflict with each other. Therefore, according to the POJG, information granules should strike a sound balance between these two aspects. An optimal information granule is the one 
for which the product of coverage and specificity criteria attains maximum.
Nowadays, the POJG has been extensively applied to construct information granules for various practical purposes, such as decision making 
\cite{QinJindong2023,ZhangBowen2023,QIN2023101833}, time series 
analysis \cite{LuWei20151,DuSheng2022,ShiWen2023,Ouyang2024} and
classification \cite{Nguyen2019,ZhuXiubin2023}.

\section{The Proposed Method}\label{Section3}
In this section, we begin by defining metrics to assess the quality of a GB according to the POJG. Subsequently, we introduce a binary tree pruning-based strategy to determine the best combination of sub-GBs for a given GB, thereby establishing the set of generated GBs. Additionally, the method for detecting abnormal GBs is presented to mitigate their impact. Finally, we outline the complete GB generation algorithm and analyze its time complexity.

\subsection{Measuring the Quality of GBs}\label{Section3.1}
According to the POJG, previous studies assess the quality of an information granule by computing the product of its coverage and its specificity \cite{QinJindong2023,LuWei20151,DuSheng2022,ShiWen2023}. Specifically, the coverage of an information granule is determined by an increasing real function of the number of instances within the information granule, while the specificity is determined by a decreasing real function of the size of the information granule. Given that each GB can be viewed as an information granule, we adopt a similar approach to define the quality of a GB in this section.

\begin{defn}\label{Definition_GB_Quality}
Let $\Omega_{X}$ be a GB derived from the dataset $X$. 
Then the coverage of $\Omega_{X}$
is given by 
\begin{align}\label{Eq_Justifiability_Degree}
  D_{C}\left(\Omega_{X}\right)=f_{1}\left(\left\vert\left\{ x\in X:\left\Vert x-C_{X}\right\Vert\leq R_{X}^{\text{Ave}}\right\}\right\vert\right),
\end{align} 
where $f_{1}$ is an increasing function. The specificity of $\Omega_{X}$
is given by 
\begin{align}\label{Eq_Specificity_Degree}
  D_{S}\left(\Omega_{X}\right)=f_{2}\left(R_{X}^{\text{Ave}}\right),
\end{align} 
where $f_{2}$ is a decreasing function. Further, the quality of $\Omega_{X}$ 
is defined as the product of coverage and specificity:
\begin{align}\label{Eq_Justifiable_Granularity_Degree}
  D\left(\Omega_{X}\right)=D_{C}\left(\Omega_{X}\right)\cdot D_{S}\left(\Omega_{X}\right).
\end{align}
\end{defn}

Evidently, Eqs. (\ref{Eq_Justifiability_Degree})~-~(\ref{Eq_Justifiable_Granularity_Degree}) provide a comprehensive approach to assessing the quality of a GB. By considering both coverage and specificity, we evaluate the GB from multiple perspectives rather than relying solely on a single criterion. This allows for a more nuanced understanding of the GB's suitability for representing the underlying data.

In the following discussions, we adopt the functions $f_{1}(t)=t$ and $f_{2}(t)=\exp(-\gamma\cdot t)$  as specified in \cite{PEDRYCZ20134209,LuWei20151}. Here, $\gamma\geq 0$ signifies the impact of specificity on the design of the information granule. This parameter offers flexibility in assessing the quality of GBs. It's worth noting that this method is independent of instance labels, making it applicable for measuring the quality of GBs in clustering tasks.

\subsection{Binary tree pruning-based strategy to generate GBs}\label{Section3.2}

In this section, we present the binary tree pruning-based strategy to generate GBs based on the quality measure of a GB. 

To expedite the generation of GBs, previous methods reviewed in subsection \ref{Section2.1} employ a top-down approach to partition GBs derived from the original dataset. This iterative process applies a specific $2$-division method until each resulting GB meets predefined criteria. While $2$-means clustering offers lower computational overhead compared to the $2$-division method outlined by Eqs. (\ref{Eq_ArgMaxXX}) and (\ref{Eq_XAlpha_XBeta}), the inherent randomness of $2$-means clustering poses the risk of producing suboptimal results, resulting in GBs that may not accurately reflect the underlying data distribution. Consequently, in this study, we adopt the method described by Eqs. (\ref{Eq_ArgMaxXX}) and (\ref{Eq_XAlpha_XBeta}) for GB partitioning, prioritizing consistency and alignment with the data distribution.

Additionally, to address the limitations of previous GB generation methods that fail to ensure the overall quality of the generated GBs, a new strategy must be developed. This strategy should guarantee that the overall quality of generated GBs surpasses that achieved by threshold-based methods or greedy strategies. The basic idea of the proposed strategy is to cease the division of a GB under two conditions: (1) when the number of instances assigned to this GB is sufficiently small, ensuring representation of instances at a coarse granularity, and (2) when further division of this GB would not enhance the overall quality of the GBs. Unlike previous strategies, the second condition is designed to maximize the overall quality of the generated GBs. We will illustrate the rationale behind the second condition using the following example.

\begin{exmp}
Let $\{\Omega_{X_{1}},\Omega_{X_{2}},\cdots,\Omega_{X_{5}}\}$ be a set of  GBs, where $\Omega_{X_{2}}$ and $\Omega_{X_{3}}$ are sub-GBs of $\Omega_{X_{1}}$;
$\Omega_{X_{4}}$ and $\Omega_{X_{5}}$ are sub-GBs of $\Omega_{X_{2}}$; $\Omega_{X_{3}},$ $\Omega_{X_{4}}$ and $\Omega_{X_{5}}$ are small enough.
Suppose that the quality levels of $\Omega_{X_{1}}$, $\Omega_{X_{2}},$ $\Omega_{X_{3}},$ $\Omega_{X_{4}}$ and $\Omega_{X_{5}}$ are $12,$ $5,$ $6,$ $4$ and $4,$ respectively. Using the greedy strategy, since the quality level of $\Omega_{X_{1}}$ exceeds the sum of the quality levels of $\Omega_{X_{2}}$ and $\Omega_{X_{3}},$ $\Omega_{X_{1}}$ is chosen to represent instances. In this scenario, the total quality level of the generated GBs is 12. However, if we continue to divide the GBs and select $\Omega_{3},$ $\Omega_{4}$, and $\Omega_{5}$ as generated GBs, then the total quality level increases to 14. Compared to the greedy strategy, continuously dividing the GBs can enhance the overall quality of the generated GBs.
\end{exmp}

To generate GBs that maximize the overall quality, we introduce the concept of the best quality level of a GB as follows.

\begin{defn}\label{Definition_Best_Quality}

Let $\Omega_{X}$ be a GB. If the number of instances belonging to $\Omega_{X}$ is sufficiently small, i.e., less than or equal to a threshold, then the best quality level of $\Omega_{X}$, denoted by $\mathcal{BQ}\left(\Omega_{X}\right)$, is defined as:
\begin{align*}
\mathcal{BQ}\left(\Omega_{X}\right) = D(\Omega_{X}),
\end{align*}
where $D(\Omega_{X})$ is calculated using Eqs. (\ref{Eq_Justifiability_Degree})-(\ref{Eq_Justifiable_Granularity_Degree}). Otherwise, if the number of instances is not small enough, the best quality level of $\Omega_{X}$ is defined as:
\begin{align}\label{Eq_BestQuality_Not_LeafNode}
\mathcal{BQ}(\Omega_{X})=\max\left(D\left(\Omega_{X}\right),
\mathcal{BQ}\left(\Omega_{X_{\alpha}}\right)+
\mathcal{BQ}\left(\Omega_{X_{\beta}}\right)\right),
\end{align}
where $X_{\alpha}$ and $X_{\beta}$ are determined by Eqs. (\ref{Eq_ArgMaxXX}) and (\ref{Eq_XAlpha_XBeta}), respectively.
\end{defn}

\begin{rem}
In Definition \ref{Definition_Best_Quality}, it is necessary to determine the threshold of the number of instances belonging to a GB. 
In Cheng et al.'s method \cite{ChengDongDong2023}, the threshold is set as $\sqrt{\vert U\vert}$. However, this value may lack flexibility in certain scenarios. Therefore, we introduce a parameter $\delta\in]0,1]$ and set the threshold as $\delta\cdot\sqrt{\vert U\vert}$.
\end{rem}

Definition \ref{Definition_Best_Quality} is evidently a recursive definition. It iteratively divides GBs, starting from the GB derived from the original dataset, until the number of instances belonging to each GB is sufficiently small to ensure representation at a coarse granularity. Subsequently, for any GB encountered in this process, its best quality level is determined by evaluating its quality level and the sum of the best quality levels of its sub-GBs. If a GB's best quality level equals its quality level, it does not require further division; otherwise, it needs to be divided. To identify the sub-GBs corresponding to a GB's best quality level, we introduce the best combination of sub-GBs for a GB as follows.

\begin{defn}\label{Definition_Best_Sub_GB}
Let $\Omega_{X}$ be a GB. If $\mathcal{BQ}(\Omega_{X})$ equals
$D(\Omega_{X}),$ then the best combination of sub-GBs of $\Omega_{X}$,
denoted by
$\mathcal{BC}(\Omega_{X}),$ is defined as
\begin{align}\label{Eq_BestCombination_LeafNode}
\mathcal{BC}(\Omega_{X})=\{\Omega_{X}\}.
\end{align}
Otherwise,
$\mathcal{BC}(\Omega_{X})$ is defined as
\begin{align}\label{Eq_BestCombination_Not_LeafNode}
\mathcal{BC}(\Omega_{X})=\mathcal{BC}(\Omega_{X_{\alpha}})\cup\mathcal{BC}(\Omega_{X_{\beta}}).
\end{align}
\end{defn}

As Definition \ref{Definition_Best_Sub_GB} relies on Definition \ref{Definition_Best_Quality}, it is also a recursive definition.

Based on the above analysis, to determine the GBs generated from the dataset $U,$ it only needs to calculate
the best combination of sub-GBs of $\Omega_{U}.$  Given that GBs are divided using Eqs. (\ref{Eq_ArgMaxXX}) and (\ref{Eq_XAlpha_XBeta}), which constitute a $2$-division method, we employ a binary tree to achieve this goal. The detailed steps of the calculation method are listed as follows:
\begin{description}
\item[\bf Step 1]: Initialize an empty tree $\mathcal{T}$ and set the GB $\Omega_{U}$ as the root node of $\mathcal{T}$.
\item[\bf Step 2]: For any node $\Omega_{X}$ in $\mathcal{T}$, if $\vert X\vert$ exceeds the threshold $\delta\cdot\sqrt{\vert U\vert}$ and $\Omega_{X}$ is a leaf node, then divide $\Omega_{X}$ into $\Omega_{X_{\alpha}}$ and $\Omega_{X_{\beta}}$ using Eqs. (\ref{Eq_ArgMaxXX}) and (\ref{Eq_XAlpha_XBeta}). Set $\Omega_{X}$ as the parent node of $\Omega_{X_{\alpha}}$ and $\Omega_{X_{\beta}}$.
\item[\bf Step 3]: Repeat Step 2 until the number of instances belonging to each leaf node of $\mathcal{T}$ is smaller than or equal to the threshold.
\item[\bf Step 4]: Calculate the quality of each node in $\mathcal{T}$ using Eqs. (\ref{Eq_Justifiability_Degree})-(\ref{Eq_Justifiable_Granularity_Degree}). Initialize the best combination of sub-GBs of each node in $\mathcal{T}$ using Eq. (\ref{Eq_BestCombination_LeafNode}).
\item[\bf Step 5]: For any leaf node $\Omega_{A}$ in $\mathcal{T}$, if its sibling node $\Omega_{B}$ is also a leaf node, then calculate the best quality level and best combination of sub-GBs of their parent node $\Omega_{C}$ using Eqs. (\ref{Eq_BestQuality_Not_LeafNode}) and (\ref{Eq_BestCombination_Not_LeafNode}). Prune leaf nodes $\Omega_{A}$ and $\Omega_{B}$ from $\mathcal{T}$.
\item[\bf Step 6]: Repeat Step 5 until only the root node $\Omega_{U}$ remains in the tree $\mathcal{T}$.
\end{description}

\begin{rem}
In the above procedure, Steps 1-3 construct a specific binary tree $\mathcal{T}$ from the dataset $U$, where each node of $\mathcal{T}$ represents a GB. Notably, each node of $\mathcal{T}$, except for the root node $\Omega_{U}$, has only one sibling node. For computational convenience, Steps 5-6 employ a bottom-up pruning strategy to ensure that the judgment conditions remain unchanged during calculation. It is evident that employing the proposed strategy for GB generation is significantly superior to using threshold-based methods or greedy strategies. 
\end{rem}

\subsection{Detecting Abnormal Granular-Balls}\label{Section3.3}

To enhance the quality of the generated GBs, it is essential to incorporate anomaly detection in the GB generation process. While the approach proposed in \cite{XieJiang2023} relies on a single distance-related indicator to identify abnormal GBs, we redefine abnormal GBs based on the POJG. A GB is deemed abnormal if both its coverage and specificity deviate from the average value.

\begin{defn}\label{Definition_Abnormal_GB}
  Let $\Phi=\{\Omega_{X_{1}},\Omega_{X_{2}},\cdots,\Omega_{X_{t}}\}$
  be a set of GBs. Denote
  \begin{align}\label{Eq_Abnormal_Detection}
    r_{\Phi}^{\text{Ave}}=
  \frac{1}{t}\sum_{i=1}^{t}R_{X_{i}}^{\text{Ave}}~\text{and}~
  \mathcal{N}_{\Phi}=\frac{1}{t}\sum_{i=1}^{t}\vert X_{i}\vert,
  \end{align}
  For any $\Omega_{X_{i}}\in\Phi,$ if $R_{X_{i}}^{\text{Ave}}>2\cdot r_{\Phi}^{\text{Ave}}$ and $\vert X_{i}\vert<\frac{1}{2}\cdot\mathcal{N}_{\Phi},$ then $\Omega_{X_{i}}$ is considered an abnormal GB.
\end{defn}

Subsequently, for a set of GBs, we can continuously divide abnormal GBs according to Definition \ref{Definition_Abnormal_GB} and Eqs. (\ref{Eq_ArgMaxXX}) and (\ref{Eq_XAlpha_XBeta}) until there are no abnormal GBs.

\subsection{Algorithm and Its Time Complexity Analysis}

\begin{figure}[h]
  \centering
  \includegraphics[width=3in]{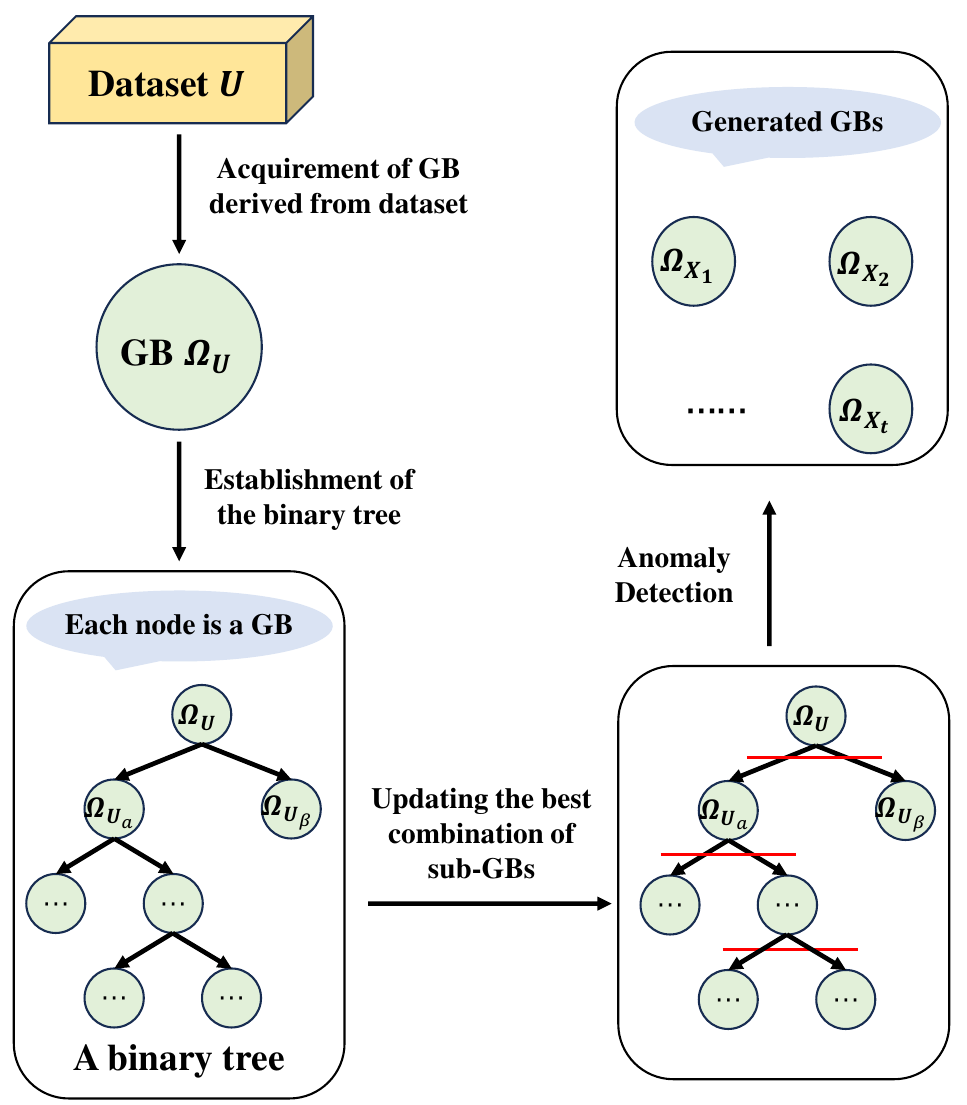}
  \caption{The framework of the proposed method.}
  \label{Fig_Framework}
\end{figure}

Based on the binary tree pruning-based strategy and the abnormal GB detection method, we propose a novel method for generating GBs for clustering tasks. The proposed generation method and its framework are illustrated in Algorithm \ref{Algorithm:3} and Fig. \ref{Fig_Framework}, respectively. The essence of Algorithm \ref{Algorithm:3} is as follows: First, it acquires a GB $\Omega_U$ from the original dataset $U$. Subsequently, the binary tree pruning-based strategy is employed to determine the generated GBs (refer to subsection \ref{Section3.2}) by updating the best combination of
sub-GBs, thereby maximizing the overall quality of generated GBs. Lastly, to enhance the rationality of the generated GBs, Algorithm \ref{Algorithm:3} incorporates anomaly detection on the generated GBs, identifying and partitioning abnormal GBs.

\begin{algorithm2e}[htbp]
  \SetAlgoLined			
  \DontPrintSemicolon		
      \SetKwInOut{Input}{\textbf{Input}}		
      \SetKwInOut{Output}{\textbf{Output}}	
  \Input{The dataset $U,$ the parameters $\gamma\in[0,+\infty[$ and $\delta\in]0,1].$}
  \Output{The set of GBs $\Phi.$}
  \textbf{initialize:} the queue $\mathcal{Q}\gets\emptyset$ 
  and the tree $\mathcal{T}\gets\emptyset;$\\ 
  add $\Omega_{U}$ to $\mathcal{Q}$ and set $\Omega_{U}$ as the root node of $\mathcal{T};$ \\
  \While{$\mathcal{Q}$ is not empty}{
    $\Omega_{X}\gets$ the first element of $\mathcal{Q};$\\
    delete the first element of $\mathcal{Q};$\\
    \If{$\vert X\vert>\delta\cdot\sqrt{\vert U\vert}$}
    {\textbf{calculate:} $X_{\alpha}$ and $X_{\beta}$ by Eqs. (\ref{Eq_ArgMaxXX}) and (\ref{Eq_XAlpha_XBeta});\\
    append $\Omega_{X_{\alpha}}$ and $\Omega_{X_{\beta}}$ to the end of $\mathcal{Q};$\\
    set $\Omega_{X_{\alpha}}$ and $\Omega_{X_{\beta}}$ as the child nodes of $\Omega_{X};$ 
    }
  }
  \ForEach{$\Omega_{X}\in \mathcal{T}$}{\textbf{initialize:} $\mathcal{BQ}(\Omega_{X})\gets D(\Omega_{X})$ by Eqs. 
  (\ref{Eq_Justifiability_Degree})-(\ref{Eq_Justifiable_Granularity_Degree}) and
   $\mathcal{BC}(\Omega_{X})\gets\{\Omega_{X}\};$}
  add all leaf nodes of $\mathcal{T}$ to $\mathcal{Q};$\\
  \While{the number of elements in $\mathcal{Q}$ is not equal to $1$}
  {$\Omega_{A}\gets$ the first element of $\mathcal{Q};$\\
  delete the first element of $\mathcal{Q};$\\
  \eIf{$\Omega_{A}$ has been deleted from $\mathcal{T}$}
  {\textbf{continue};}
  {$\Omega_{B}\gets$ the sibling node of $\Omega_{A}$;\\
  $\Omega_{C}\gets$ the parent node of $\Omega_{A}$;}
  \eIf{$\Omega_{B}$ is not a leaf node}
  {append $\Omega_{A}$ into the end of $\mathcal{Q};$\\
  \textbf{continue};}
  {delete $\Omega_{A}$ and $\Omega_{B}$ from $\mathcal{T};$\\ 
  append $\Omega_{C}$ to the end of $\mathcal{Q};$}
  \If{$\mathcal{BQ}(\Omega_{A})+\mathcal{BQ}(\Omega_{B})\geq\mathcal{BQ}(\Omega_{C})$}
  {\textbf{update:} $\mathcal{BQ}(\Omega_{C})\gets\mathcal{BQ}(\Omega_{A})+\mathcal{BQ}(\Omega_{B})$
  and $\mathcal{BC}(\Omega_{C})\gets \mathcal{BC}(\Omega_{A})\cup \mathcal{BC}(\Omega_{B});$}
  }
  \textbf{initialize:} the queue $\mathcal{Q}\gets\emptyset$ and the set of GBs $\Phi\gets\emptyset;$\\
  add all elements in $\mathcal{BC}(\Omega_{U})$ to $\mathcal{Q};$\\
  \textbf{calculate:} $r_{\Phi}^{\text{Ave}}$ and $\mathcal{N}_{\Phi}$ by Eq. (\ref{Eq_Abnormal_Detection});\\
  \While{$\mathcal{Q}$ is not empty}
  { $\Omega_{X}\gets$ the first element of $\mathcal{Q};$\\
  delete the first element of $\mathcal{Q};$\\
    \eIf{$R_{X}^{\text{Ave}}>2\cdot r_{\Phi}^{\text{Ave}}$
    and $\vert X\vert<\frac{1}{2}\cdot\mathcal{N}_{\Phi},$}
    {\textbf{calculate:} $X_{\alpha}$ and $X_{\beta}$ by Eqs. (\ref{Eq_ArgMaxXX}) and (\ref{Eq_XAlpha_XBeta});\\
   append $\Omega_{X_{\alpha}}$ and $\Omega_{X_{\beta}}$ to the end of $\mathcal{Q};$}
    {$\Phi\gets\Phi\cup\{\Omega_{X}\};$}}
  \textbf{return:} the set of GBs $\Phi.$
  \caption{The proposed GB generation method}
  \label{Algorithm:3}
\end{algorithm2e}

We now analyze the time complexity of Algorithm \ref{Algorithm:3}. Let us suppose that the number of instances is denoted by $n$, and after line 11 of Algorithm \ref{Algorithm:3}, the number of generated GBs is represented by $k$, where $k\lll n$. As Algorithm \ref{Algorithm:3} adopts the same GB division approach as Xie et al.'s method \cite{XieJiang2023}, the time complexity of steps 3~-~11 is approximately $O(n\log n)$, consistent with the complexity of Xie et al.'s method \cite{XieJiang2023}. Step 13 has a time complexity of approximately $O(n)$. Consequently, the time complexity of steps 12-14 is approximately $O(kn)$. Steps 15~-~35 have a time complexity of $O(k)$, while steps 37~-~48 also have a complexity of $O(k)$. Thus, the overall time complexity of Algorithm \ref{Algorithm:3} is approximately $O(n\log n + (2+n)k)$

\section{Numerical Experiments}\label{Section4}

In this section, we demonstrate the performance of the proposed GB generation method through several numerical experiments. These experiments were conducted using MATLAB 2023b on a computer equipped with 32.0 GB of RAM, an Intel Core i9-13900 HX CPU running at 2.20 GHz, and running Windows 11.
\subsection{Clustering on Synthetic Datasets}\label{Section_Synthetic_Datasets}

We first compare the proposed GB generation method with two existing methods proposed by \cite{ChengDongDong2023} and \cite{XieJiang2023} on six synthetic datasets. Specifically, each GB generation method is incorporated into the GBSC algorithm \cite{XieJiang2023} to generate clustering results for different datasets. Details about the used synthetic datasets are provided in Table \ref{Table_synthetic_datasets}. We evaluate the three methods based on three criteria: (1) the number of GBs generated from the dataset; (2) the clustering results obtained using GBSC; and (3) the computing overhead of GBSC.

\begin{table}[htbp]
  \centering
  \fontsize{9}{9}\selectfont
  \begin{threeparttable}
  \caption{Description of synthetic datasets}
  \label{Table_synthetic_datasets}
    \begin{tabular}{cccccc}
    \toprule
     Datasets & Sources & Instances & Dimensions & Classes\\
    \midrule
    SD1 & \cite{Rodriguez2014} & 788  & 2 & 7\\
    SD2 & \cite{XieJiang2023} & 1039  & 2 & 4\\
    SD3 & \cite{XieJiang2023} & 1052  & 2 & 2\\
    SD4 & \cite{XieJiang2023} & 1741  & 2 & 6\\
    SD5 & \cite{XieJiang2023} & 1016  & 2 & 4\\
    SD6 & \cite{ChengDongDong2018}& 6700  & 2 & 5\\
    \bottomrule
    \end{tabular}
    \end{threeparttable}
\end{table}

\begin{figure*}[htbp]
  \centering
  \subfigure[]{\includegraphics[width=1in,height=1in]{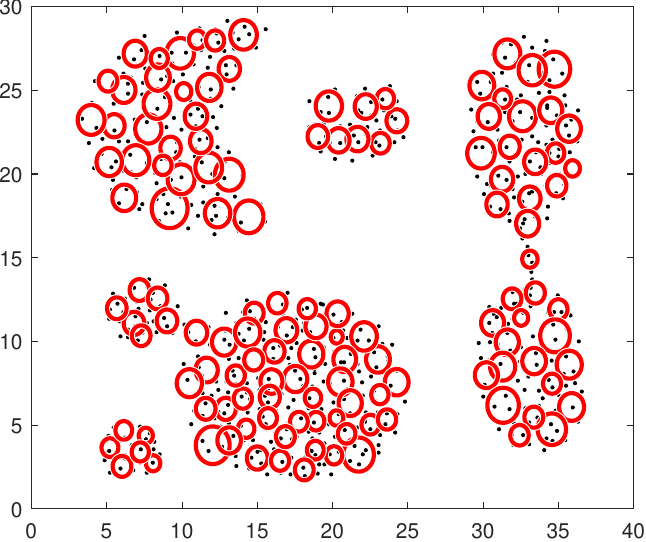}}
  \subfigure[]{\includegraphics[width=1in,height=1in]{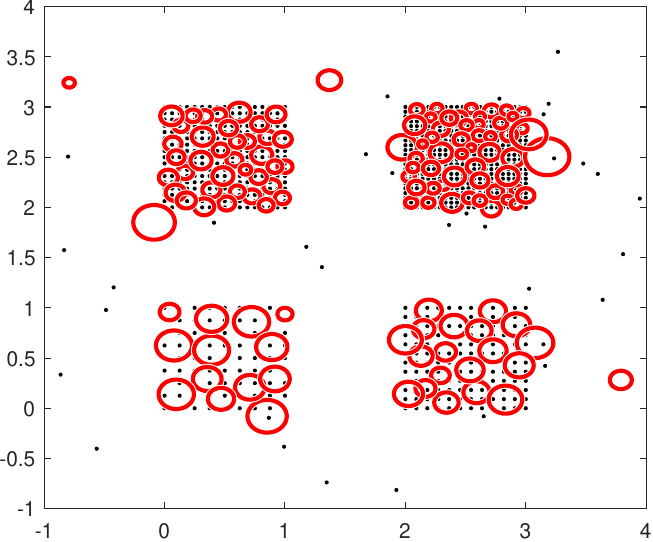}}
  \subfigure[]{\includegraphics[width=1in,height=1in]{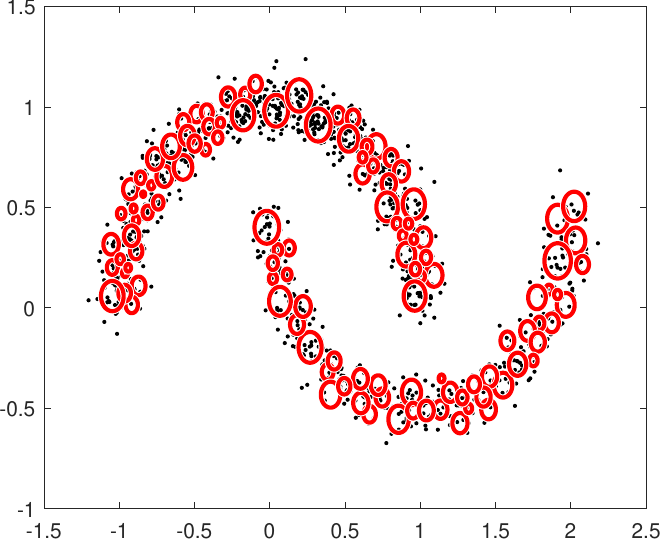}}
  \subfigure[]{\includegraphics[width=1in,height=1in]{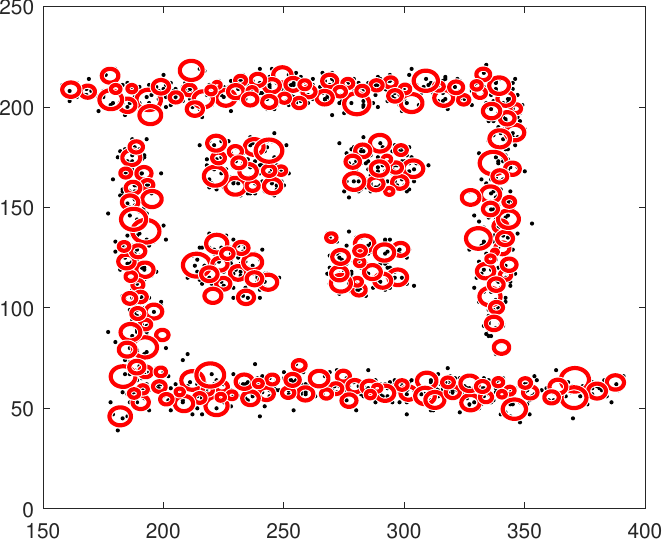}}
  \subfigure[]{\includegraphics[width=1in,height=1in]{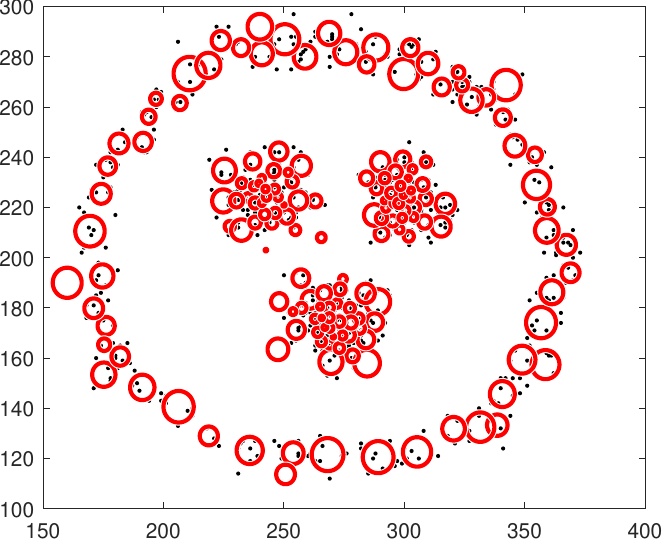}}
  \subfigure[]{\includegraphics[width=1in,height=1in]{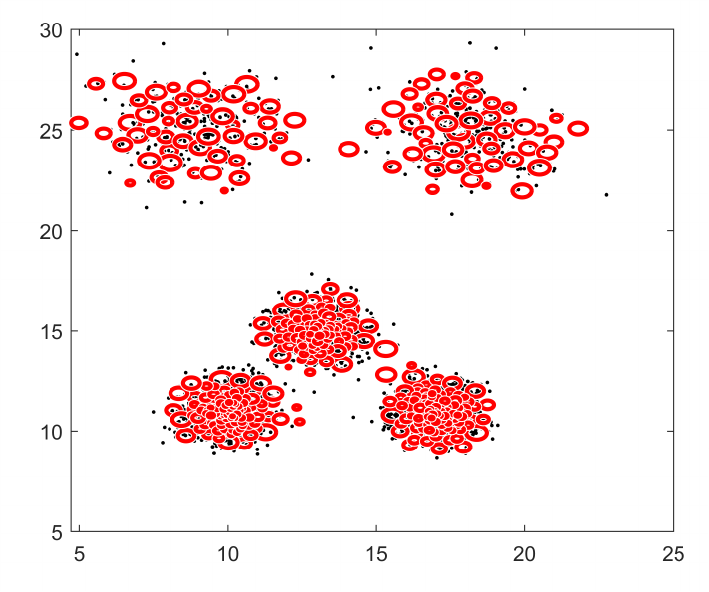}}
  \subfigure[]{\includegraphics[width=1in,height=1in]{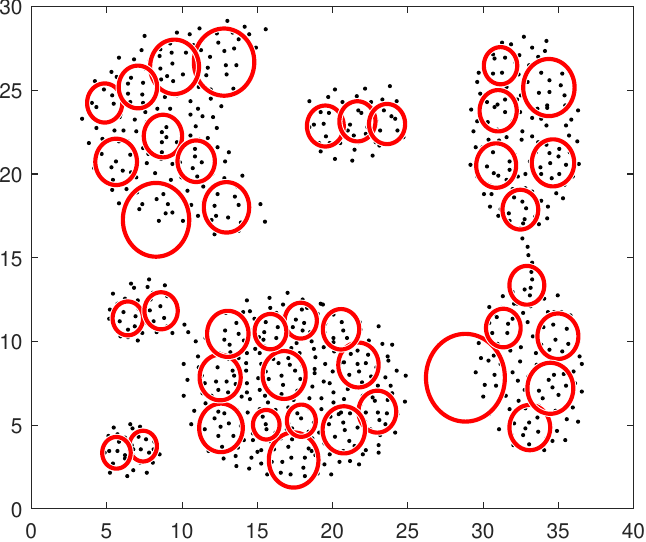}\label{Fig_NGB_Cheng_1}}
  \subfigure[]{\includegraphics[width=1in,height=1in]{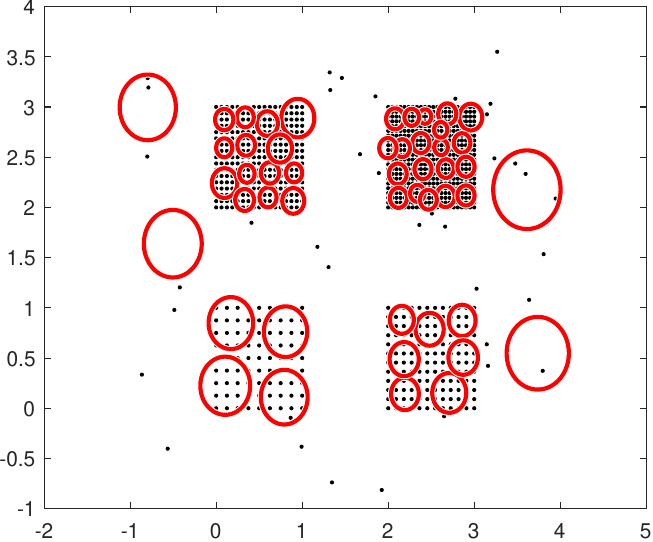}}
  \subfigure[]{\includegraphics[width=1in,height=1in]{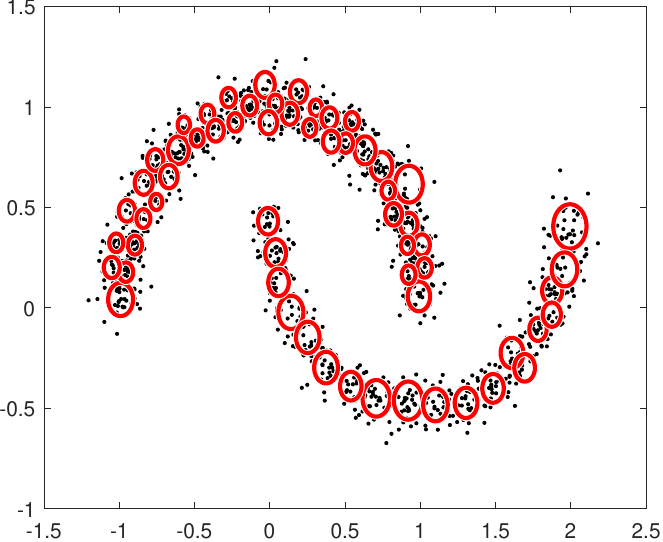}}
  \subfigure[]{\includegraphics[width=1in,height=1in]{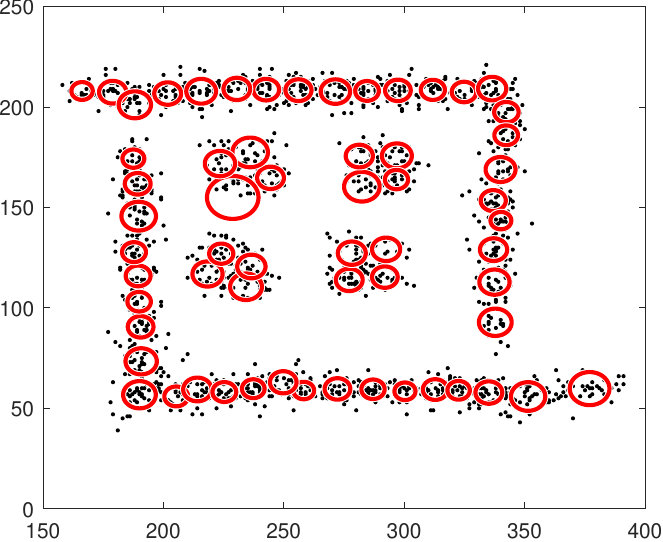}}
  \subfigure[]{\includegraphics[width=1in,height=1in]{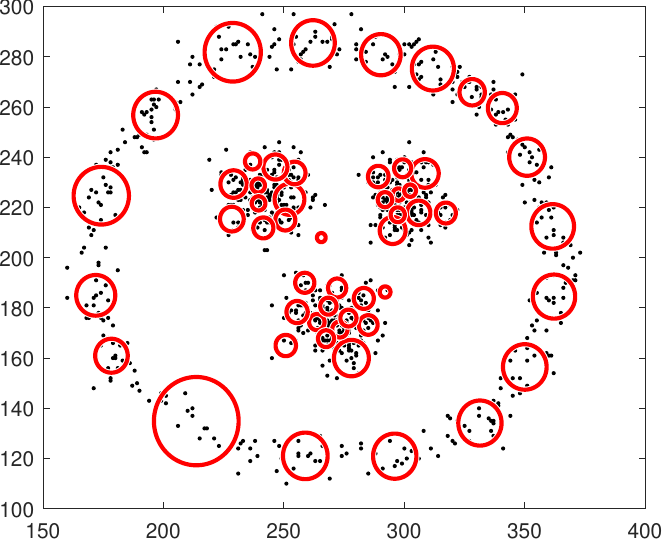}\label{Fig_NGB_Cheng_5}}
  \subfigure[]{\includegraphics[width=1in,height=1in]{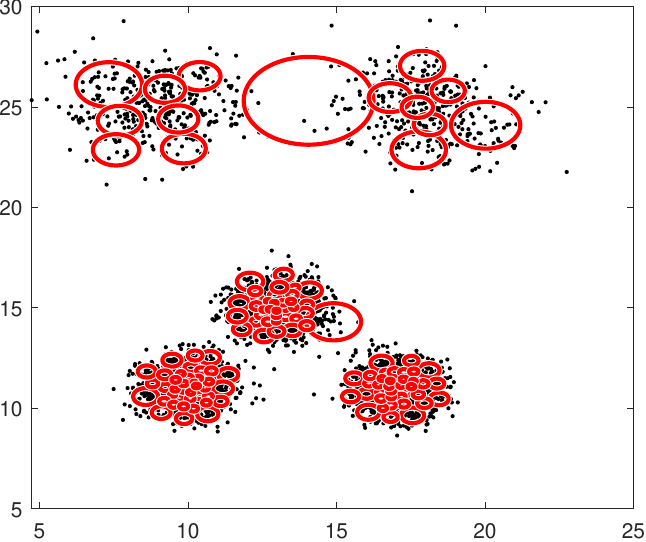}\label{Fig_NGB_Cheng_6}}
  \subfigure[]{\includegraphics[width=1in,height=1in]{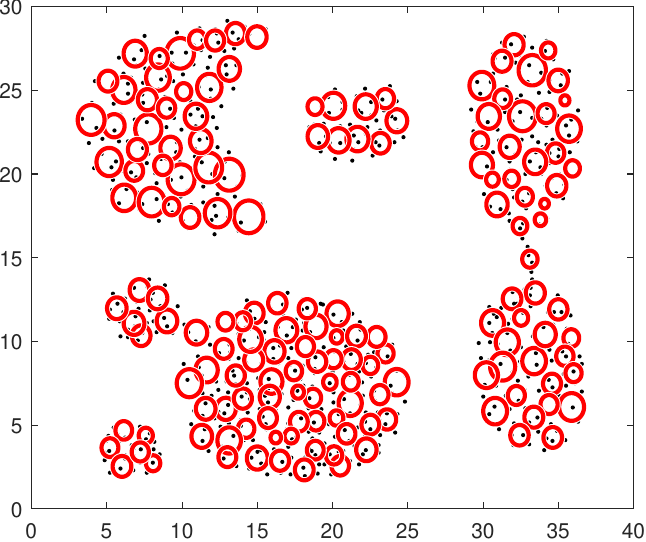}}
  \subfigure[]{\includegraphics[width=1in,height=1in]{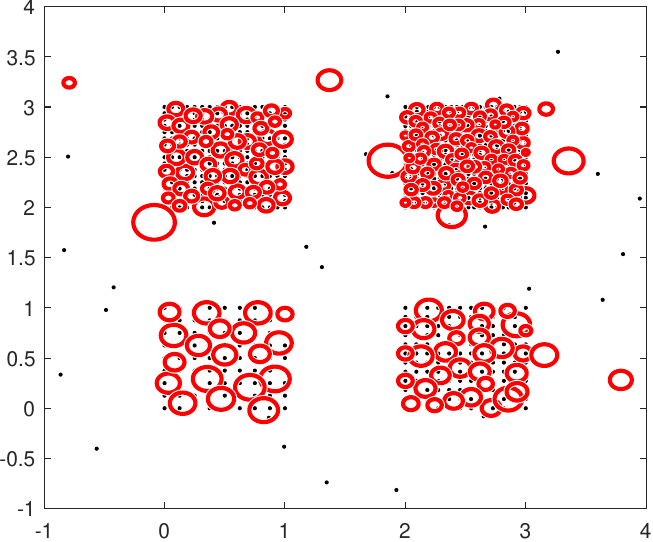}}
  \subfigure[]{\includegraphics[width=1in,height=1in]{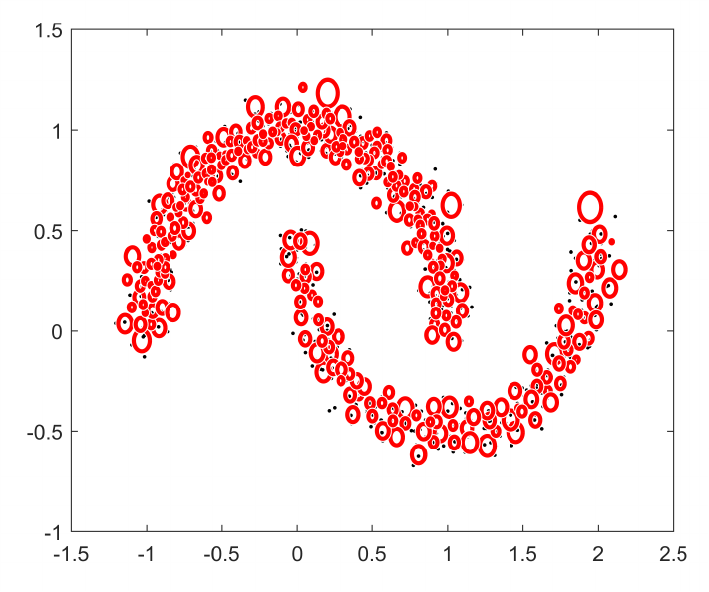}}
  \subfigure[]{\includegraphics[width=1in,height=1in]{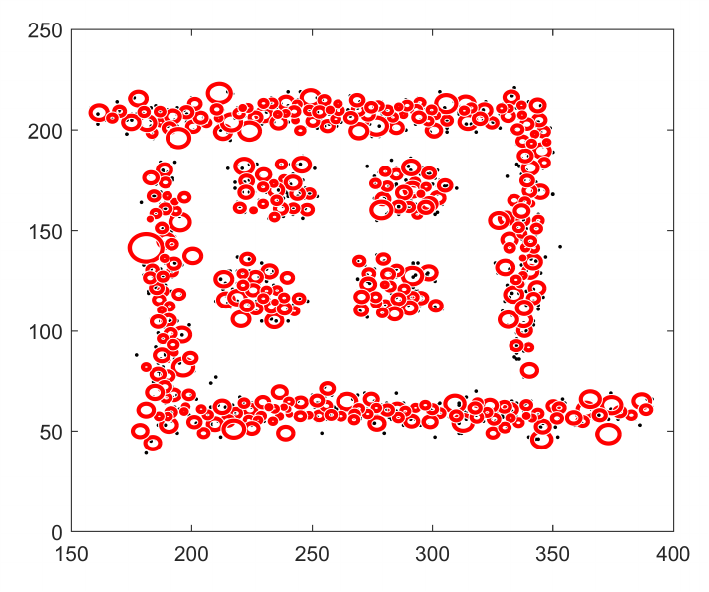}}
  \subfigure[]{\includegraphics[width=1in,height=1in]{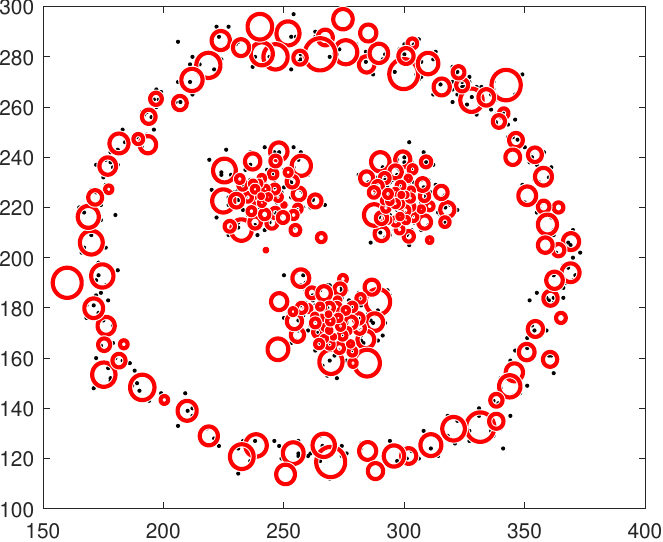}}
  \subfigure[]{\includegraphics[width=1in,height=1in]{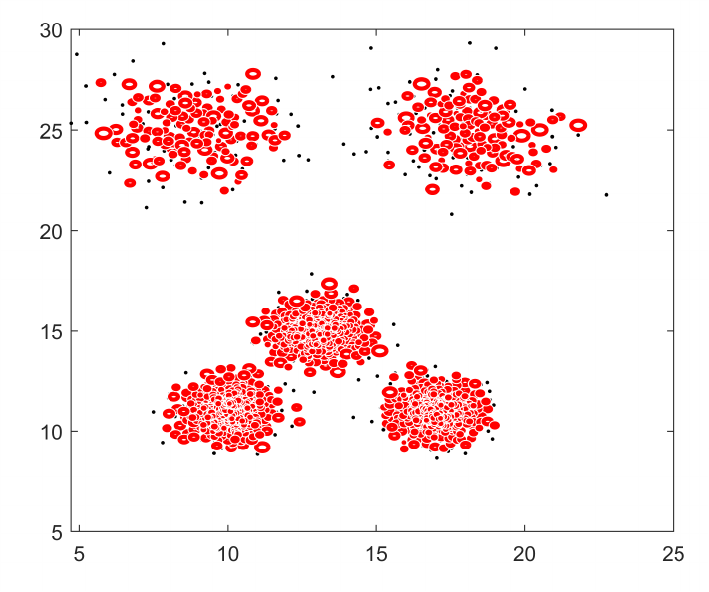}}
  \caption{The GBs with average radius generated from the six synthetic datasets. 
  (a)-(f) Our method. 
  (g)-(l) Cheng et al.'s method \cite{ChengDongDong2023}.
  (m)-(r) Xie et al.'s method \cite{XieJiang2023}.
  }
  \label{Fig_Number_Granular_Balls}
\end{figure*}
\begin{figure*}[htbp]
  \centering
  \subfigure[]{\includegraphics[width=1in,height=1in]{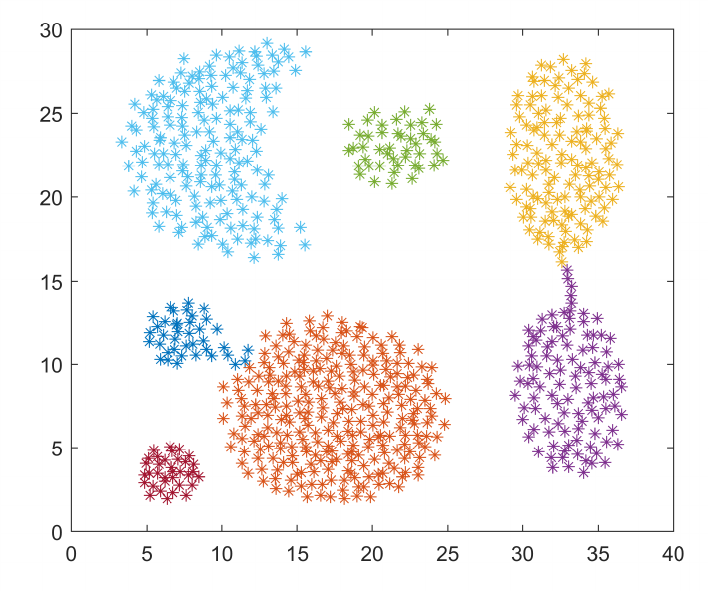}}
  \subfigure[]{\includegraphics[width=1in,height=1in]{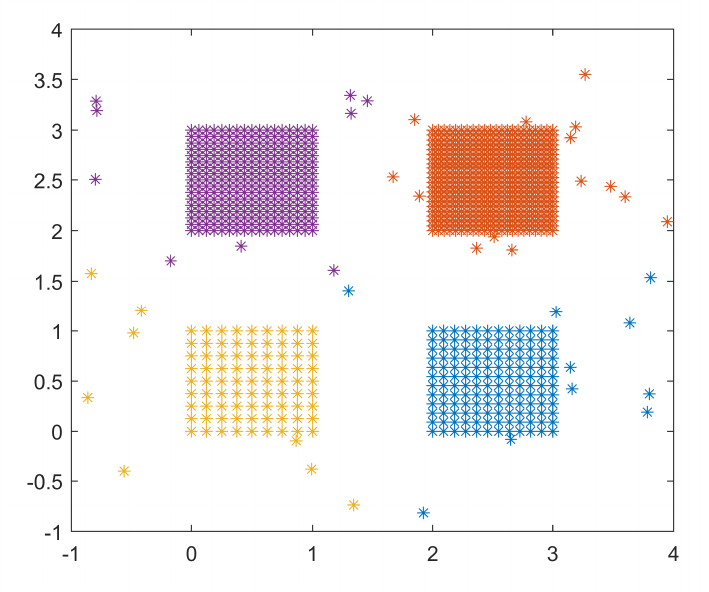}}
  \subfigure[]{\includegraphics[width=1in,height=1in]{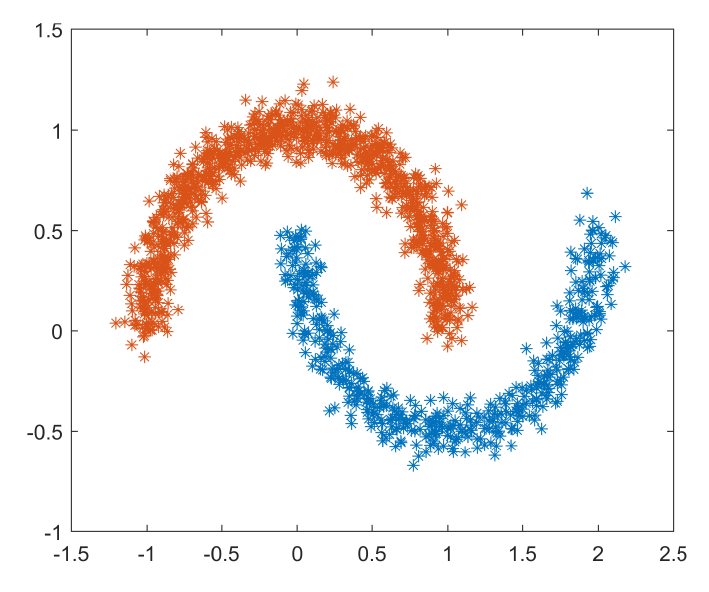}}
  \subfigure[]{\includegraphics[width=1in,height=1in]{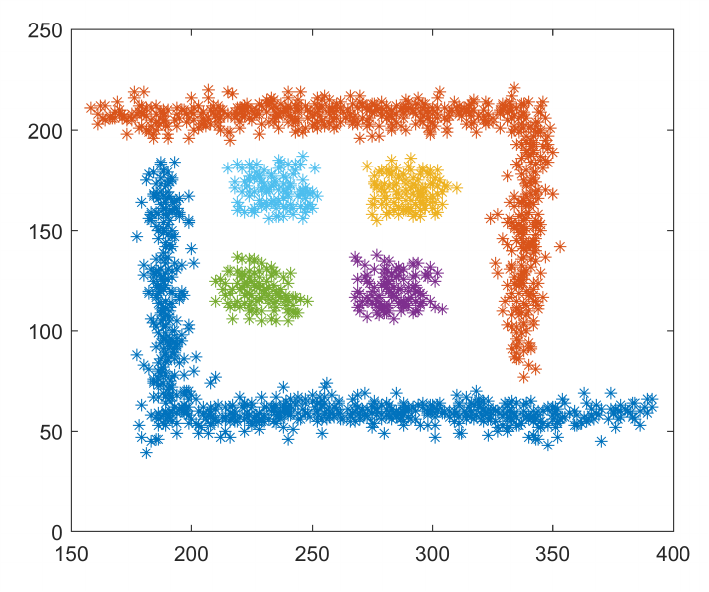}}
  \subfigure[]{\includegraphics[width=1in,height=1in]{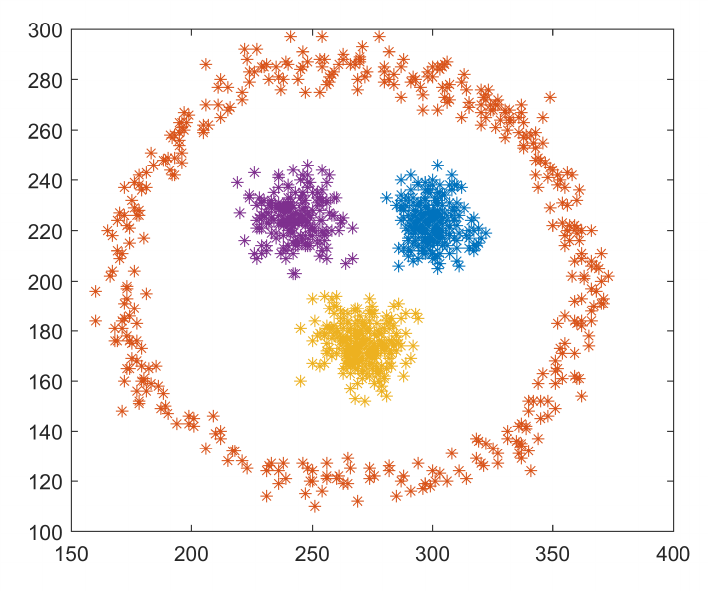}}
  \subfigure[]{\includegraphics[width=1in,height=1in]{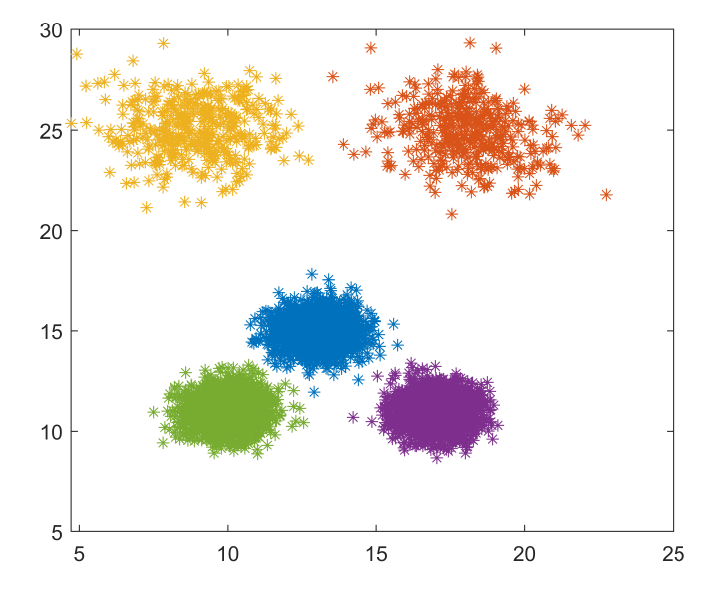}}
  \subfigure[]{\includegraphics[width=1in,height=1in]{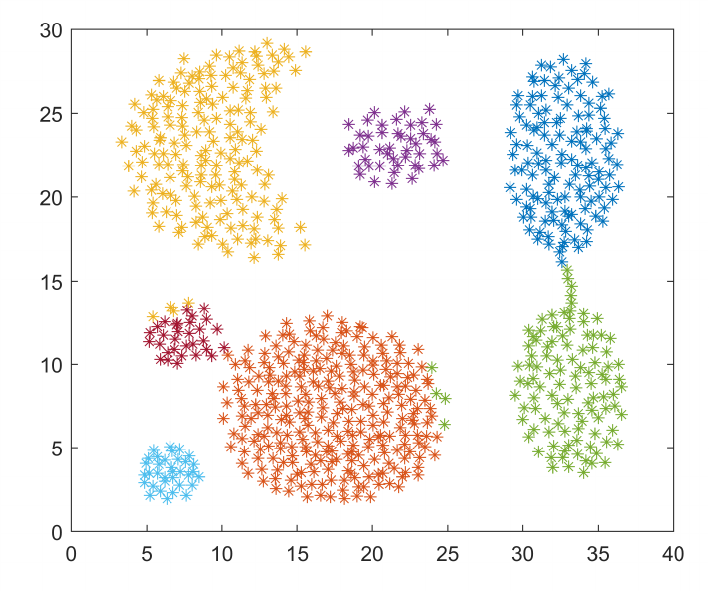}\label{Fig_Cheng_Clustering_1}}
  \subfigure[]{\includegraphics[width=1in,height=1in]{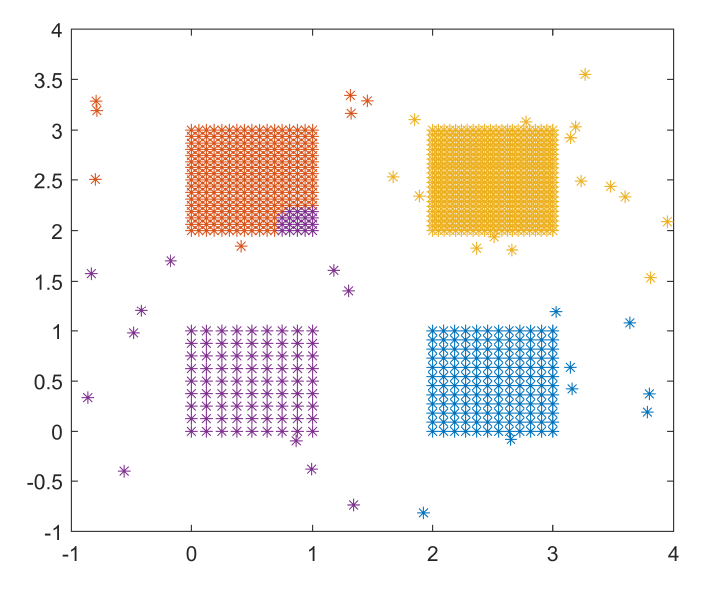}}
  \subfigure[]{\includegraphics[width=1in,height=1in]{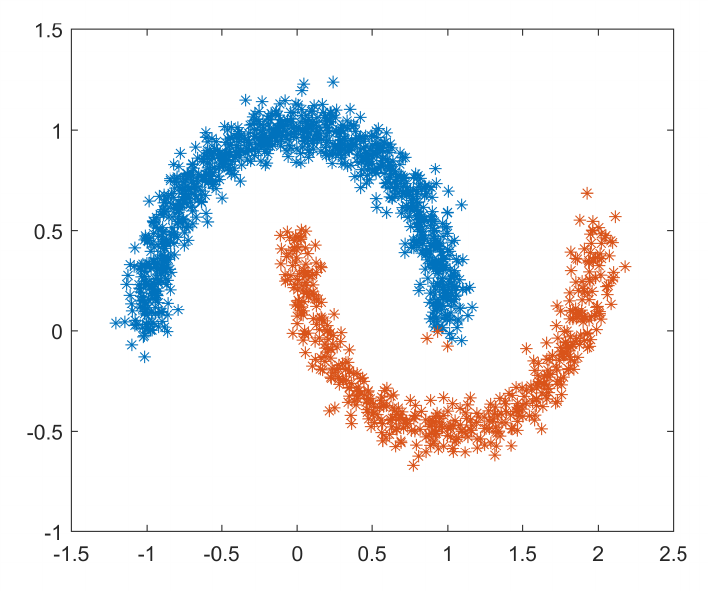}}
  \subfigure[]{\includegraphics[width=1in,height=1in]{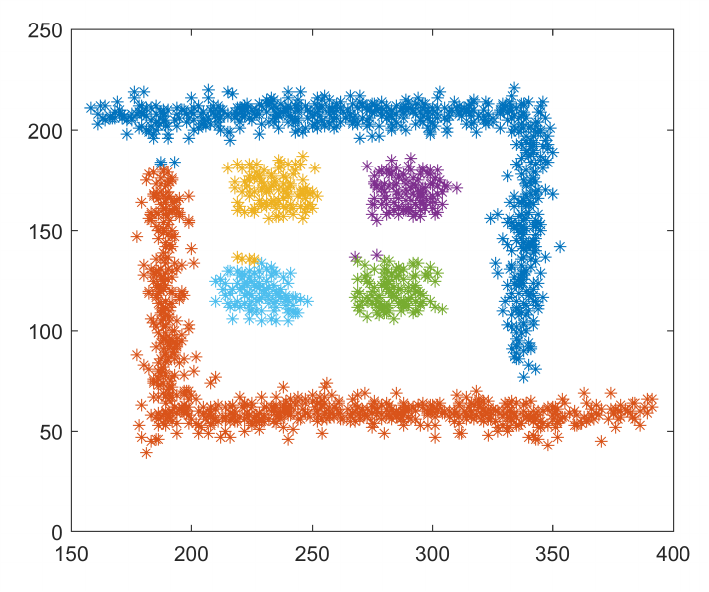}}
  \subfigure[]{\includegraphics[width=1in,height=1in]{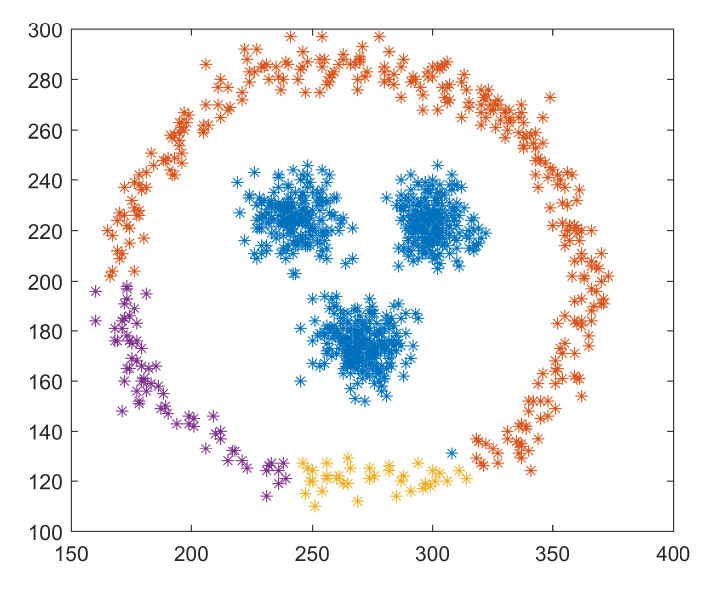}}
  \subfigure[]{\includegraphics[width=1in,height=1in]{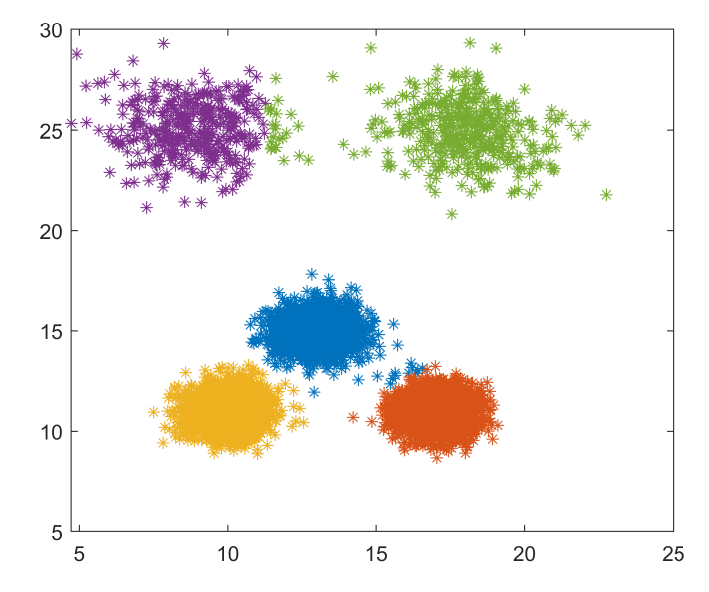}\label{Fig_Cheng_Clustering_6}}
  \subfigure[]{\includegraphics[width=1in,height=1in]{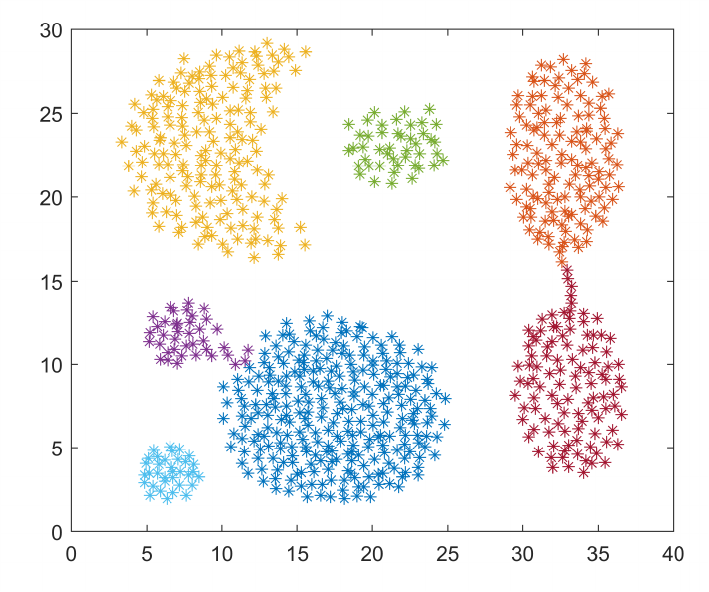}}
  \subfigure[]{\includegraphics[width=1in,height=1in]{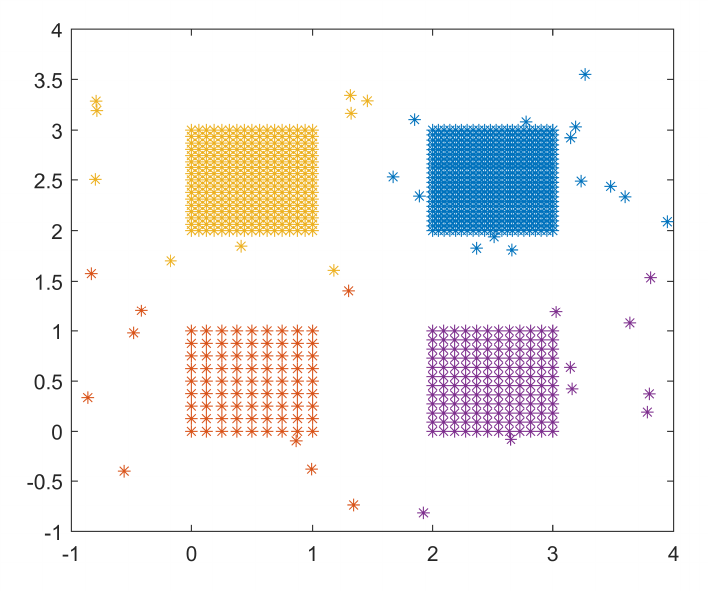}}
  \subfigure[]{\includegraphics[width=1in,height=1in]{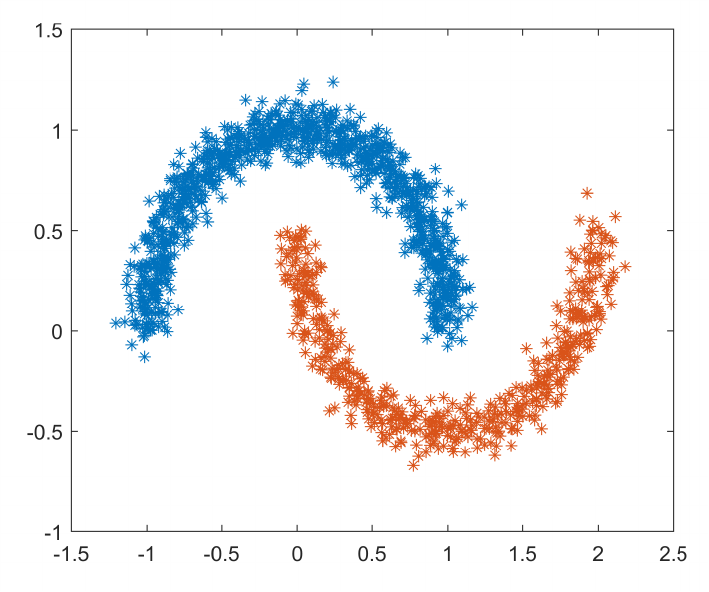}}
  \subfigure[]{\includegraphics[width=1in,height=1in]{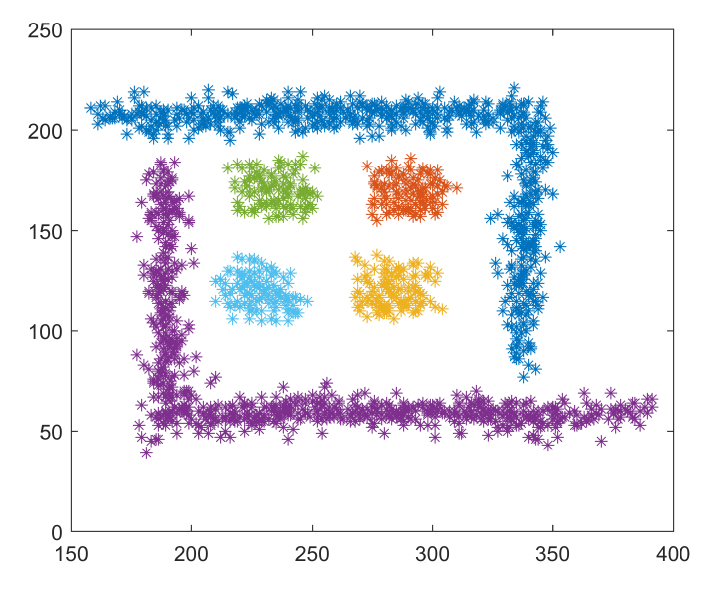}}
  \subfigure[]{\includegraphics[width=1in,height=1in]{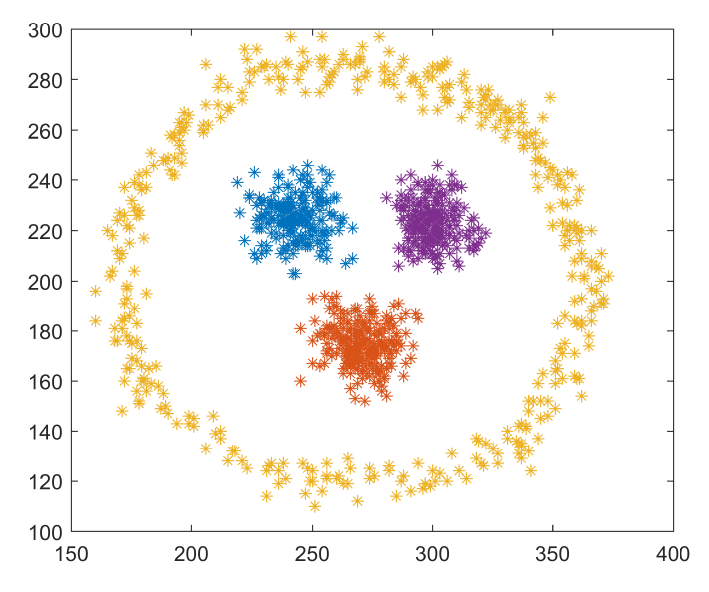}}
  \subfigure[]{\includegraphics[width=1in,height=1in]{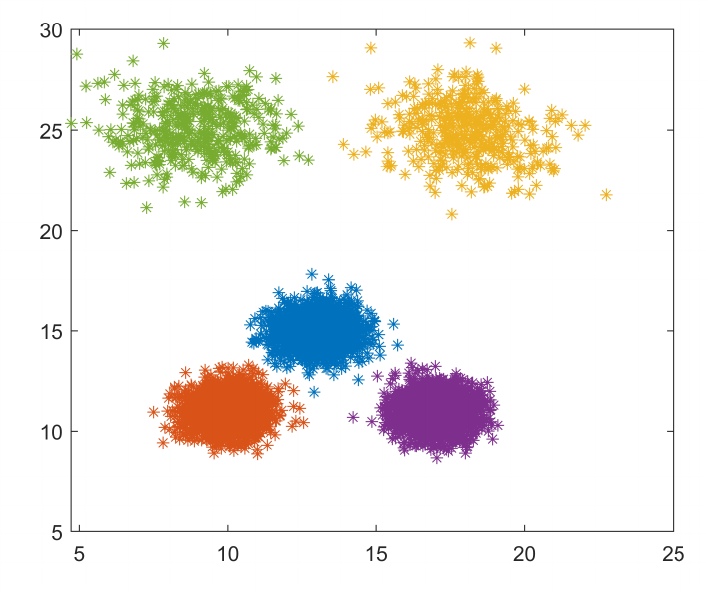}}
  \caption{The clustering results of GBSC on the six synthetic datasets. 
  (a)-(f) Our method. 
  (g)-(l) Cheng et al.'s method \cite{ChengDongDong2023}.
  (m)-(r) Xie et al.'s method \cite{XieJiang2023}.
  }
  \label{Fig_Classification_Result}
\end{figure*}
\begin{table}[htbp]
  \centering
  \fontsize{8}{8}\selectfont
  \renewcommand\tabcolsep{4pt}
  \begin{threeparttable}
  \caption{The number of generated GBs and the computing overhead (in seconds) of GBSC on six synthetic datasets}
  \label{Table_Running_Time_Number_GB}
    \begin{tabular}{ccccccc}
    \toprule
    \multirow{2}{*}{Datasets} & \multicolumn{2}{c}{Our Method} & \multicolumn{2}{c}{Cheng et al.'s Method}        & \multicolumn{2}{c}{Xie et al.'s Method}          \\
    \cmidrule(lr){2-3}\cmidrule(lr){4-5}\cmidrule(lr){6-7}
    & \multicolumn{1}{c}{NGGB} & \multicolumn{1}{c}{CO} & \multicolumn{1}{c}{NGGB} & \multicolumn{1}{c}{CO} & \multicolumn{1}{c}{NGGB} & \multicolumn{1}{c}{CO} \\
    \midrule
SD1                       &          133              &     0.109                    &        41                 &        0.138                 &    161                    &0.109\\
SD2                       &          145              &     0.126                    &        50                 &        0.161                 &    228                    &0.160\\
SD3                       &          127              &     0.128                    &        60                 &        0.149                 &    329                    &0.157\\
SD4                       &          227              &     0.177                    &        61                 &        0.150                 &    389                    &0.204\\
SD5                       &          170              &     0.148                    &        52                 &        0.150                 &    229                    &0.128\\
SD6                       &          436              &     0.594                    &        127                &        0.221                 &    1607                   &1.523\\
    \bottomrule
    \end{tabular}
    Note: ``NGGB'' and ``CO'' denote the number of generated GBs and
     the computing overhead, respectively.
    \end{threeparttable}
\end{table}
\begin{figure*}[htbp]
  \centering
  \subfigure[]{\includegraphics[width=3in,height=2in]{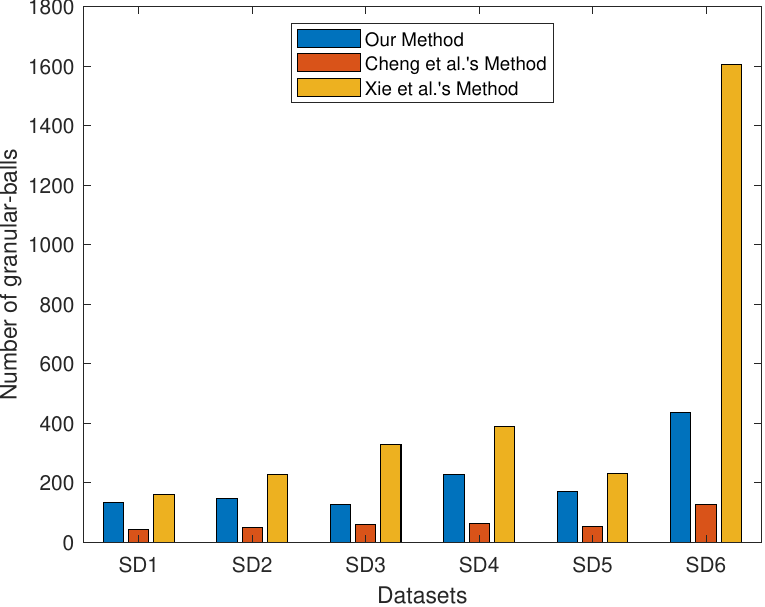}\label{Fig_NGB_synthetic_dataset}}
  \subfigure[]{\includegraphics[width=3in,height=2in]{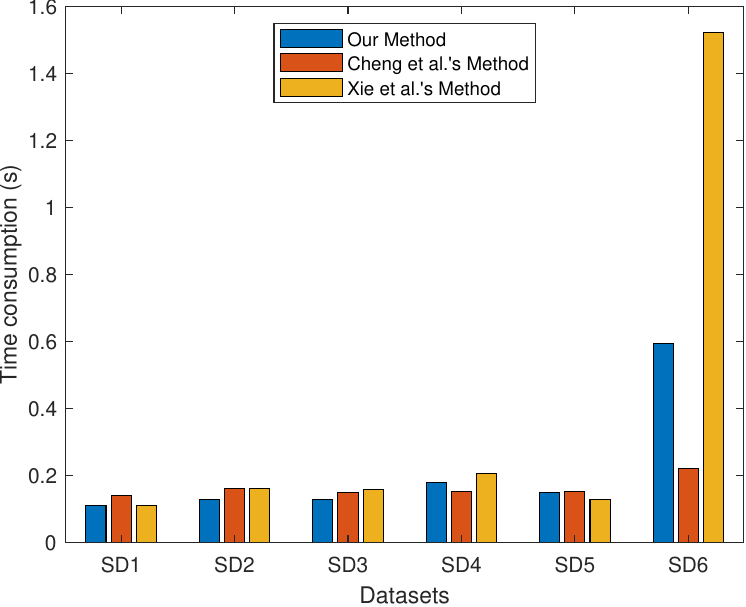}\label{Fig_TC_synthetic_dataset}}
  \caption{Comparison of three generation methods of GBs on synthetic datasets. 
  (a) The number of generated GBs.
  (b) The computing overhead (in seconds) of GBSC.
  }
  \label{Fig_Time_Consumption_Synthetic_Datasets}
\end{figure*}

In this subsection, the parameters $\gamma$ and $\delta$ in our method are set to $1$ and $0.3$, respectively. Meanwhile, in the GBSC algorithm \cite{XieJiang2023}, the parameter $\sigma$, used to calculate the similarity between GBs, is set to $\frac{\sqrt{2}}{2}$, $2$, $0.1$, $2$, $2$, and $2$, respectively, for synthetic datasets SD1-SD6. The selection of these parameters is to ensure that the clustering results are consistent as possible with the intuitive clustering results. The experimental results are depicted in Figs. \ref{Fig_Number_Granular_Balls}~-~\ref{Fig_Time_Consumption_Synthetic_Datasets} and summarized in Table \ref{Table_Running_Time_Number_GB}. In Fig. \ref{Fig_Number_Granular_Balls}, instances and GBs are represented in black and red, respectively, while Fig. \ref{Fig_Classification_Result} illustrates instances belonging to different clusters in various colors.

We further analyze the clustering results obtained on synthetic datasets. Firstly, from Table \ref{Table_Running_Time_Number_GB} and Figs. \ref{Fig_Number_Granular_Balls} and \ref{Fig_NGB_synthetic_dataset}, it is evident that Cheng et al.'s (resp. Xie et al.'s) method generates the lowest (resp. highest) number of GBs for each synthetic dataset. However, some GBs generated by Cheng et al.'s method exhibit large average radii (see, e.g., Figs. \ref{Fig_NGB_Cheng_1}, \ref{Fig_NGB_Cheng_5}, and \ref{Fig_NGB_Cheng_6}). Secondly, from Fig. \ref{Fig_Classification_Result}, it is clear that the clustering results of GBSC obtained using our method or Xie et al.'s method align well with intuitive clustering results. Conversely, the clustering results of GBSC derived from Cheng et al.'s method appear somewhat counterintuitive across all synthetic datasets. Particularly on synthetic datasets SD1, SD5, and SD6, the counterintuitive clustering results of GBSC from Cheng et al.'s method are evidently associated with the generated GBs having large average radii and a small number of instances. Hence, for clustering tasks, the anomaly detection of generated GBs is crucial. Thirdly, from Table \ref{Table_Running_Time_Number_GB} and Fig. \ref{Fig_TC_synthetic_dataset}, the three methods exhibit similar performance on five small-scale synthetic datasets in terms of computing overhead. However, on the large-scale synthetic dataset SD6, there is a significant disparity in the computing overhead of the three methods, with Cheng et al.'s (resp. Xie et al.'s) method showing the lowest (resp. highest) computing overhead. This ranking aligns with the ranking of the number of generated GBs.

The experimental results demonstrate that the clustering results of GBSC obtained by our method are consistent with intuitive clustering results, outperforming those derived from Cheng et al.'s method. Meanwhile, from the perspective of computing overhead, our method outperforms Xie et al.'s method on large-scale synthetic datasets.

\subsection{Clustering on Publicly Available Datasets}\label{Section_Real_Datasets}
\begin{table}[htbp]
  \centering
  \fontsize{9}{9}\selectfont
  \begin{threeparttable}
  \caption{Description of publicly available datasets}
  \label{Table_real_datasets}
    \begin{tabular}{cccccc}
    \toprule
     ID &Dataset & Instances & Dimensions & Classes\\
    \midrule
    D1&iris  & 150  & 4 & 3\\
    D2&wine  & 178  & 13 & 3\\
    D3&wpbc  & 194 & 32 & 2\\
    D4&sonar  & 208  & 60 & 2\\
    D5&seed  & 210  & 7 & 3\\
    D6&ionosphere  & 351 & 34 & 2\\
    D7&cancer  & 683 & 9 & 2\\
    D8&cardiotocography  & 2126 & 21 & 3\\
    D9&abalone  & 4177  & 8 & 3\\
    D10&waveform  & 5000  & 21 & 3\\
    D11&Electrical Grid& 10000 & 13 & 2\\
    D12&pendigits  & 10992  & 16 & 10\\
    \bottomrule
    \end{tabular}
    \end{threeparttable}
\end{table}

\begin{table}[htbp]
  \centering
  \fontsize{9}{9}\selectfont
  \begin{threeparttable}
  \caption{The parameters settings on publicly available datasets}
  \label{Table_parameters_real_dataset}
    \begin{tabular}{ccccccc}
      \toprule
      \multirow{2}{*}{ID} & \multirow{2}{*}{$\sigma$} & \multirow{2}{*}{$\lambda$} & \multicolumn{2}{c}{$\gamma$} &\multicolumn{2}{c}{$\delta$} \\
        \cmidrule(lr){4-5}\cmidrule(lr){6-7}
        ~ & ~ & ~&GBDPC & GBSC & GBDPC & GBSC \\
        \midrule
        D1 & 1 & 0.1 & 1 & 2 & 0.5 & 1 \\
        D2 & 180 & 0.3 & 1 & 8 & 0.5 & 0.4 \\
        D3 & 70 & 0.2 & 10 & 1 & 0.7 & 0.1 \\
        D4 & 2 & 0.1 & 1 & 5 & 0.6 & 0.6 \\
        D5 & 1 & 0.3 & 1 & 2 & 0.4 & 0.1 \\
        D6 & 2 & 0.5& 1 & 1 & 0.7 & 1 \\
        D7 & 4 & 0.4& 1 & 1 & 0.4 & 0.6 \\
        D8 & 2 & 0.4& 1 & 2 & 0.4 & 0.2 \\
        D9 & 9 & 0.04& 3 & 1 & 1 & 0.15 \\
        D10 & 10 & 0.4& 1 & 9 & 0.8 & 0.1 \\
        D11 & 2 & 0.1& 2 & 1 & 0.2 & 0.9 \\
        D12 & 50 & 0.1& 1 & 1 & 0.4 & 0.2 \\
        \bottomrule
    \end{tabular}
    \end{threeparttable}
\end{table}

\begin{table*}[htbp]
  \centering
  \fontsize{9}{9}\selectfont
  \renewcommand\tabcolsep{4pt}
  \begin{threeparttable}
  \caption{Clustering accuracy (\%), NMI and computing overhead (in seconds) on twelve publicly available datasets}
  \label{Table_ACC_TC_Real_Datasets}
    \begin{tabular}{ccccccccccccc}
    \toprule
    \multirow{2}{*}{Dataset} & & \multicolumn{2}{c}{Our Method} & \multicolumn{2}{c}{Cheng et al.'s Method}        & \multicolumn{2}{c}{Xie et al.'s Method}    & \multirow{2}{*}{DPC} & \multirow{2}{*}{SC} & \multirow{2}{*}{kmeans++}  &   \multirow{2}{*}{U-SPEC}&   \multirow{2}{*}{U-SENC}\\
    \cmidrule(lr){3-4}\cmidrule(lr){5-6}\cmidrule(lr){7-8}
    & & \multicolumn{1}{c}{GBDPC} & \multicolumn{1}{c}{GBSC} & \multicolumn{1}{c}{GBDPC} & \multicolumn{1}{c}{GBSC} & \multicolumn{1}{c}{GBDPC} & \multicolumn{1}{c}{GBSC}\\
    \midrule
\multirow{3}{*}{D1} 
& ACC & \textbf{96.67} & 92.67 & 73.10 & 88.25 & 88.00 & 90.00 & 95.33 & 89.59 & 86.74 & 94.67 & 74.93\\
& NMI& \textbf{0.8801}& 0.7959& 0.7069& 0.7570& 0.7561& 0.7660& 0.8572& 0.7559& 0.7331 & 0.8431& 0.6225\\
& CO & 0.0014& 0.0015& 0.0083& 0.0089& 0.0045& 0.0048& 0.0003& 0.0087& 0.0006 & 0.0100& 0.2082\\
\midrule
\multirow{3}{*}{D2} 
& ACC & \textbf{71.35} & 69.67 & 63.57 & 66.85 & 70.22 & 68.12 & 65.73 & 67.10 & 69.50 & 57.35 & 47.46\\
& NMI& 0.3883& \textbf{0.4343}& 0.3342& 0.3852& 0.3983& 0.3827& 0.4219& 0.4063& 0.4260 & 0.2542& 0.1089\\
& CO & 0.0026& 0.0058& 0.0085& 0.0092& 0.0049& 0.0067& 0.0004& 0.0121& 0.0008 & 0.0097& 0.2217\\
\midrule
\multirow{3}{*}{D3} 
& ACC& \textbf{77.32} & 76.80 & 76.29 & 76.59 & 76.29 & 76.80 & 76.29 & 76.80 & 76.29 & 76.80 & 76.29\\
& NMI& \textbf{0.0388}& 0.0257& 0.0167& 0.0249& 0.0174& 0.0257& 0.0005& 0.0257& 0.0195& 0.0136& 0.0149\\
& CO & 0.0023& 0.0086& 0.0108& 0.0119& 0.0065& 0.0074& 0.0004& 0.0193& 0.0007& 0.0122& 0.2504\\
\midrule
\multirow{3}{*}{D4} 
& ACC &\textbf{61.54} & 56.73 & 54.29 & 55.44 & 55.77 & 55.29 & 54.33 & 54.33 &54.56 & 53.36 & 54.07\\
& NMI& \textbf{0.0587}& 0.0165& 0.0174& 0.0103& 0.0464& 0.0099& 0.0028& 0.0058&0.0070& 0.0429& 0.0262\\
& CO & 0.0027& 0.0037& 0.0119& 0.0149& 0.0061& 0.0066& 0.0005& 0.0194&0.0010& 0.0131& 0.2695\\
\midrule
\multirow{3}{*}{D5} 
& ACC& 88.57 & \textbf{90.00} & 82.50 & 86.22 & 80.00 & 89.05 & 67.14 & 89.05 & 89.35 & 67.07 & 71.41\\
& NMI& 0.6868& \textbf{0.6979}& 0.6230& 0.6400& 0.6173& 0.6812& 0.5142& 0.6869& 0.7005& 0.4657& 0.5040\\
& CO & 0.0035& 0.0095& 0.0093& 0.0104& 0.0065& 0.0079& 0.0005& 0.0184& 0.0008& 0.0120& 0.2498\\
\midrule
\multirow{3}{*}{D6} 
& ACC & 67.24 & \textbf{73.22} & 64.67 & 71.53 & 64.10 & 70.37 & 66.10 & 70.79 & 70.89 & 64.39 & 69.42\\
& NMI & 0.1123& \textbf{0.1708}& 0.0475& 0.1411& 0.0705& 0.1264& 0.0278& 0.1273& 0.1293& 0.0045& 0.1071\\
& CO  & 0.0056& 0.0069& 0.0274& 0.0261& 0.0172& 0.0221& 0.0011& 0.0388& 0.0008& 0.0250& 0.5035\\
\midrule
\multirow{3}{*}{D7} 
& ACC& 68.23 & 96.78 & 66.29 & 96.52 & 65.01 & 96.63 & 73.79 & 96.49 & 96.07 & 65.15 & \textbf{96.95}\\
& NMI& 0.2890& 0.7828& 0.1695& 0.7817& 0.1554& 0.7749& 0.1799& 0.7684& 0.7491& 0.0025& \textbf{0.8017}\\
& CO & 0.0607& 0.0503& 0.0270& 0.0275& 0.0641& 0.0700& 0.0036& 0.1190& 0.0006& 0.0700& 1.8008\\
\midrule
\multirow{3}{*}{D8} 
& ACC & \textbf{80.24} & 78.56 & 78.37 & 78.13 & 77.85 & 78.43 & 77.89 & 77.85 & 77.85 & 78.49 & 77.86\\
& NMI & \textbf{0.1972}& 0.0504& 0.0942& 0.0525& 0.0486& 0.0419& 0.0031& 0.0186& 0.0697& 0.0268& 0.0846\\
& CO  & 0.0777& 0.1840& 0.0523& 0.0546& 0.1837& 0.4510& 0.0436& 5.8344& 0.0033& 0.1571& 3.3477\\
\midrule
\multirow{3}{*}{D9}
& ACC& 51.64 & 50.30 & 49.16 & 50.36 & 44.58 & 50.34 & \textbf{52.60} & 50.89 & 49.68 & 40.80 & 40.63\\
& NMI& 0.1356& 0.1152& 0.1356& 0.1231& 0.1009& 0.1179& \textbf{0.1630}& 0.1253& 0.1106& 0.0247& 0.0286\\
& CO & 0.1813& 0.4047& 0.0594& 0.0583& 0.4489& 0.6313& 0.1834& 11.323& 0.0014& 0.2350& 6.4301\\
\midrule
\multirow{3}{*}{D10} 
& ACC & \textbf{69.18} & 62.43 & 60.54 & 46.30 & 66.44 & 61.80 & 67.80 & 51.20 & 52.44 & 52.83 & 51.69\\
& NMI & \textbf{0.3973}& 0.3575& 0.3619& 0.1340& 0.3490& 0.3580& 0.2649& 0.3679& 0.3606& 0.3680& 0.3671\\
& CO  & 0.2069& 0.6817& 0.1840& 0.1238& 0.7515& 0.7584& 0.3284& 27.770& 0.0055& 0.2855& 6.6731\\
\midrule
\multirow{3}{*}{D11} 
& ACC & 64.89 & \textbf{69.97} & 64.07 & 63.94 & 63.80 & 67.68 & 63.80 & 66.95 & 63.80 & 64.73 & 64.53\\
& NMI & 0.0166& \textbf{0.1201}& 0.0219& 0.0216& 0.0060& 0.1039& 0.0152& 0.0933& 0.0316& 0.0583& 0.0522\\
& CO  & 1.1632& 0.9369& 0.1562& 0.1567& 1.6721& 2.6094& 1.2287& 331.15& 0.0089& 0.4518& 15.085\\
\midrule
\multirow{3}{*}{D12} 
& ACC & 75.79 & 72.20 & 69.62 & 68.42 & 58.26 & 71.76 & 76.73 & 71.64 & 71.73 & 82.90 & \textbf{88.29}\\
& NMI & 0.7077& 0.6757& 0.6933& 0.6384& 0.6619& 0.6810& 0.7454& 0.6874& 0.6790& 0.7955& \textbf{0.8507}\\
& CO  & 1.0900& 1.7535& 0.1713& 0.1837& 2.1632& 4.1159& 1.4295& 397.52& 0.0366& 0.5646& 9.6800\\
\midrule
\multirow{3}{*}{Average} 
& ACC & 72.72 & \textbf{74.11} & 66.87 & 70.71 & 67.53 & 73.02 & 69.79 & 71.88 & 71.58 & 66.55 & 67.79\\
& NMI & 0.3257& \textbf{0.3536}&  0.2685& 0.3091& 0.2690& 0.3391& 0.2663& 0.3392& 0.3347& 0.2417& 0.3007\\
& CO  & 0.2332& 0.3373& 0.0605& 0.0572& 0.4441& 0.7243& 0.2684& 55.315& 0.0051& 0.1538& 3.7267\\
  \bottomrule
    \end{tabular}
    Note: ``ACC'' and ``CO'' denote accuracy and
     computing overhead, respectively.
    \end{threeparttable}
\end{table*}
\begin{figure*}[htbp]
  \centering
  \subfigure[]{\includegraphics[width=3.5in, height=2in]{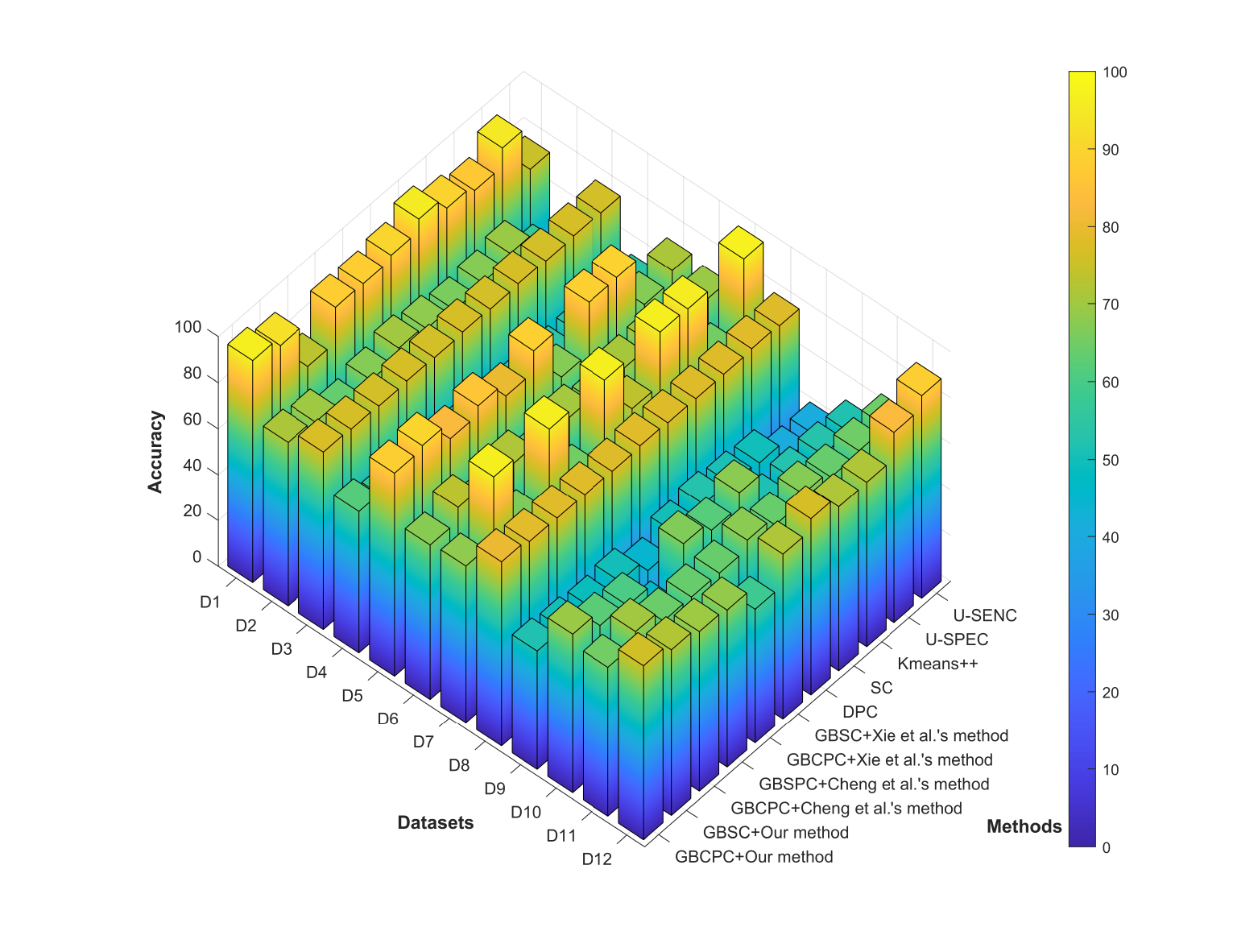}\label{Fig_RealDatasetACC}}
  \subfigure[]{\includegraphics[width=3.5in, height=2in]{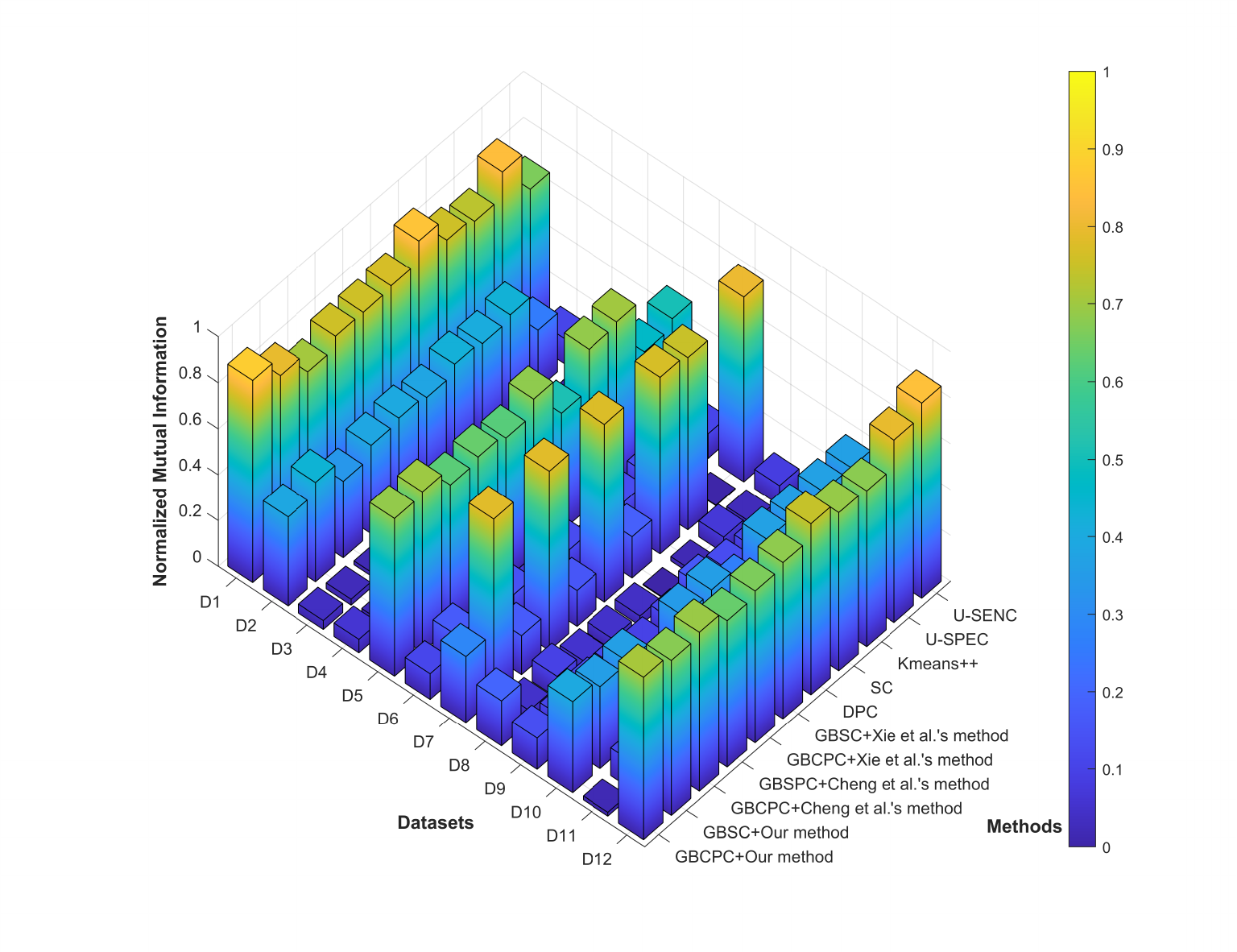}\label{Fig_RealDatasetNMI}}
  \caption{Comparison of some clustering algorithms on twelve publicly available datasets. (a) Clustering accuracy (\%). (b) Clustering NMI.}
  \label{Fig_RealDatasetACCNMI}
\end{figure*}

\begin{figure*}[htbp]
  \centering
  \subfigure[]{\includegraphics[width=1.6in]{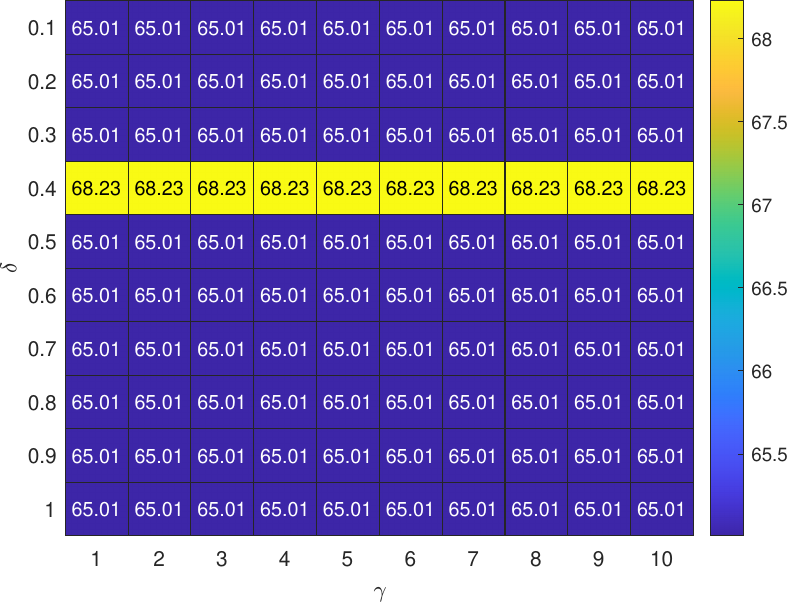}\label{Fig_parameter_analysis_ACC_D7}}
  \subfigure[]{\includegraphics[width=1.6in]{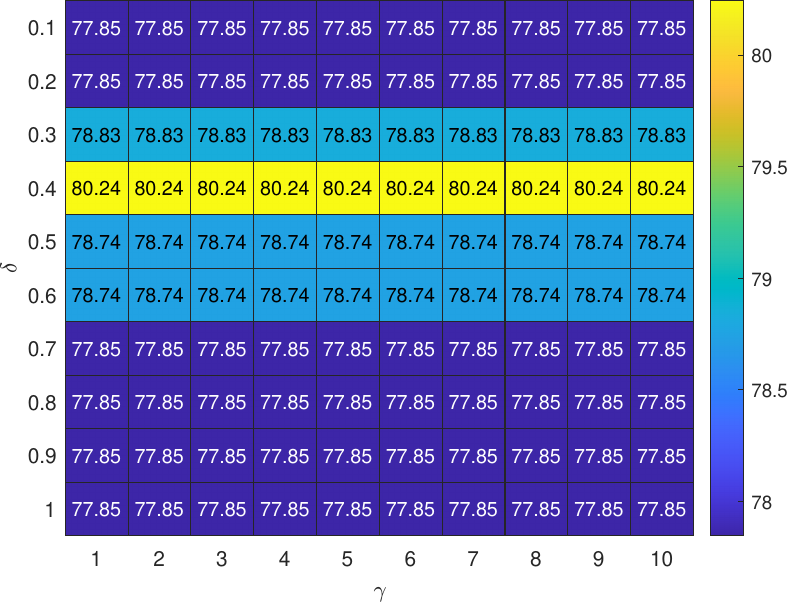}\label{Fig_parameter_analysis_ACC_D8}}
  \subfigure[]{\includegraphics[width=1.6in]{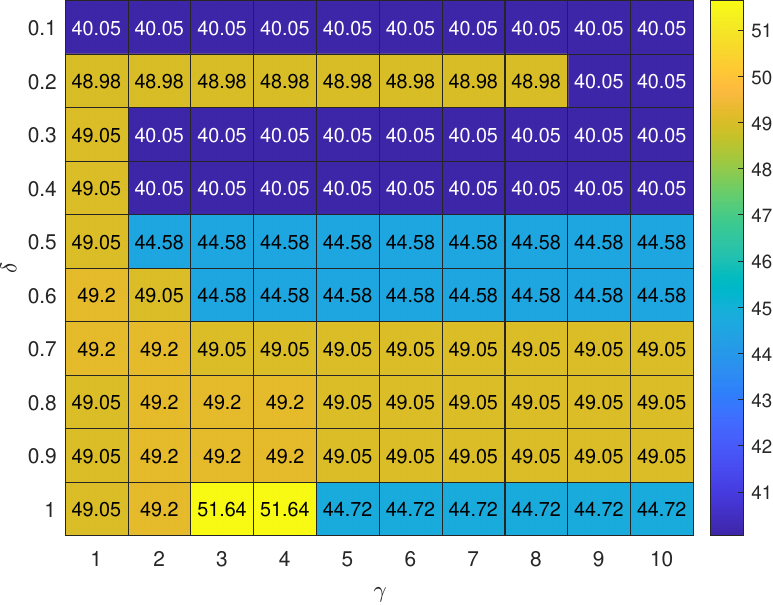}\label{Fig_parameter_analysis_ACC_D9}}
  \subfigure[]{\includegraphics[width=1.6in]{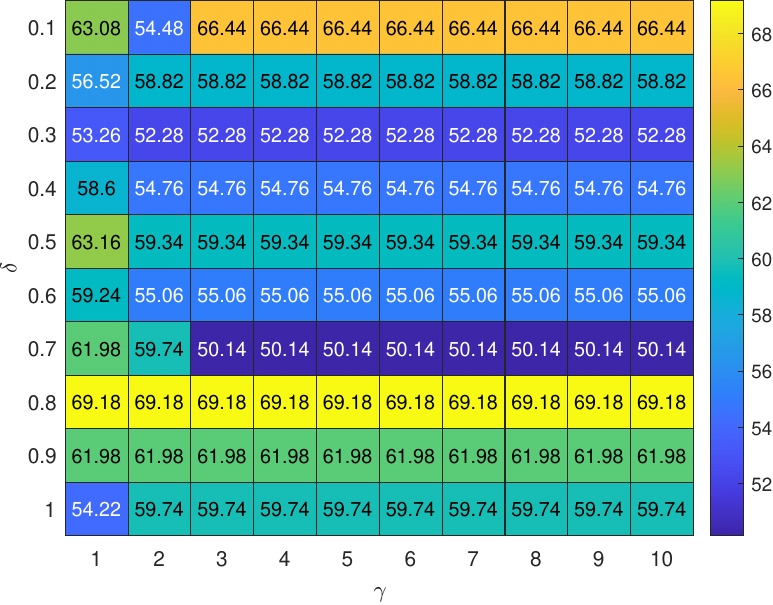}\label{Fig_parameter_analysis_ACC_D10}}
  \subfigure[]{\includegraphics[width=1.6in]{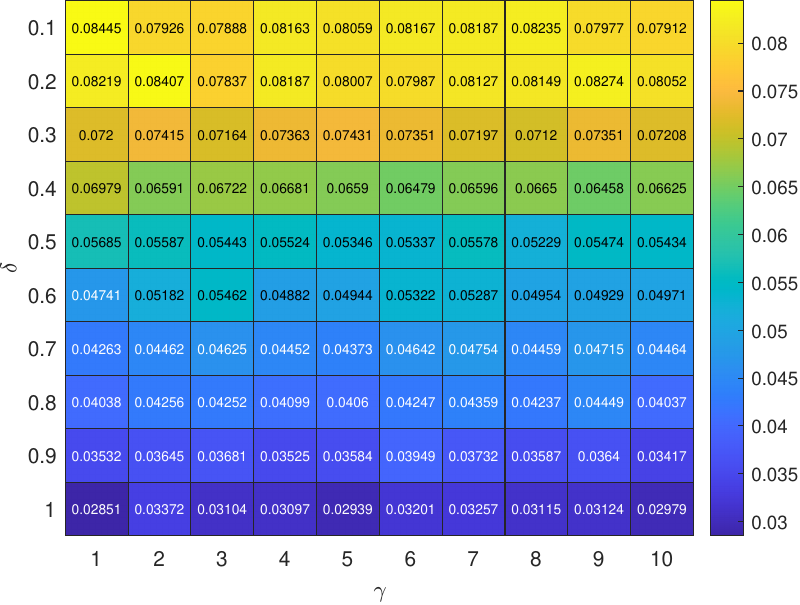}\label{Fig_parameter_analysis_TC_D7}}
  \subfigure[]{\includegraphics[width=1.6in]{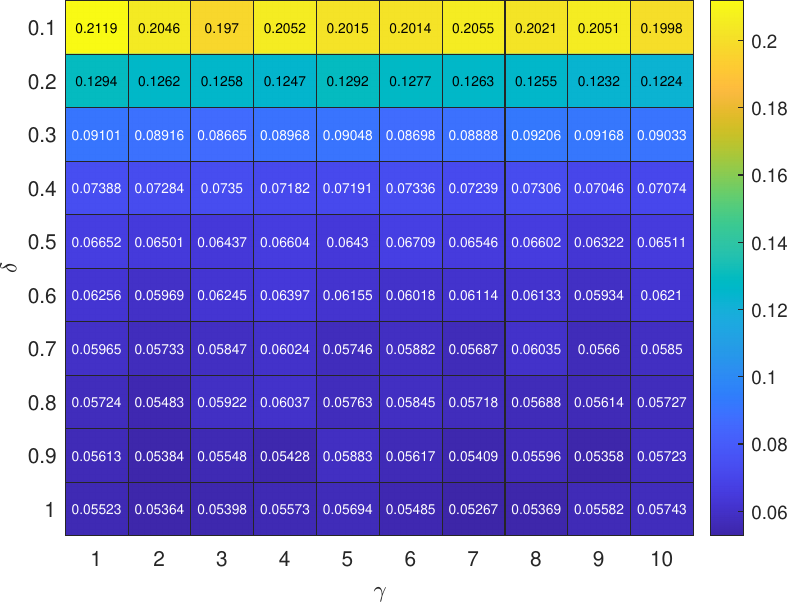}\label{Fig_parameter_analysis_TC_D8}}
  \subfigure[]{\includegraphics[width=1.6in]{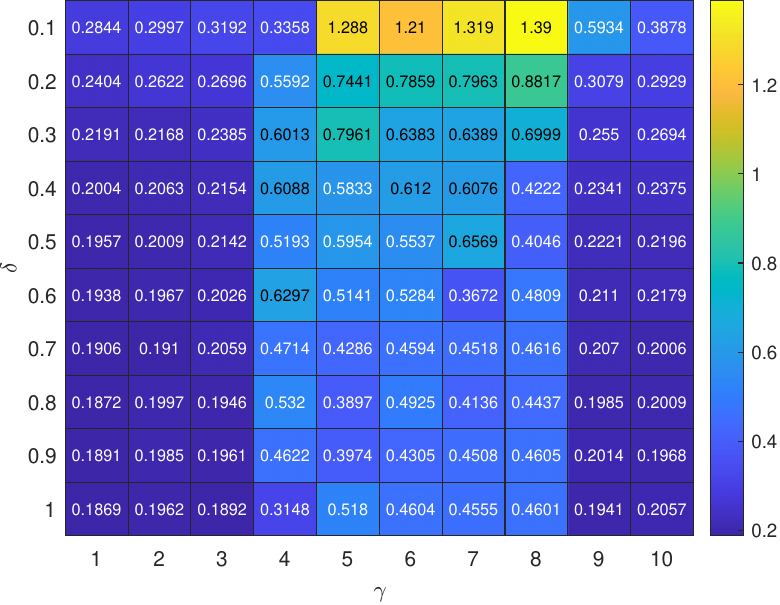}\label{Fig_parameter_analysis_TC_D9}}
  \subfigure[]{\includegraphics[width=1.6in]{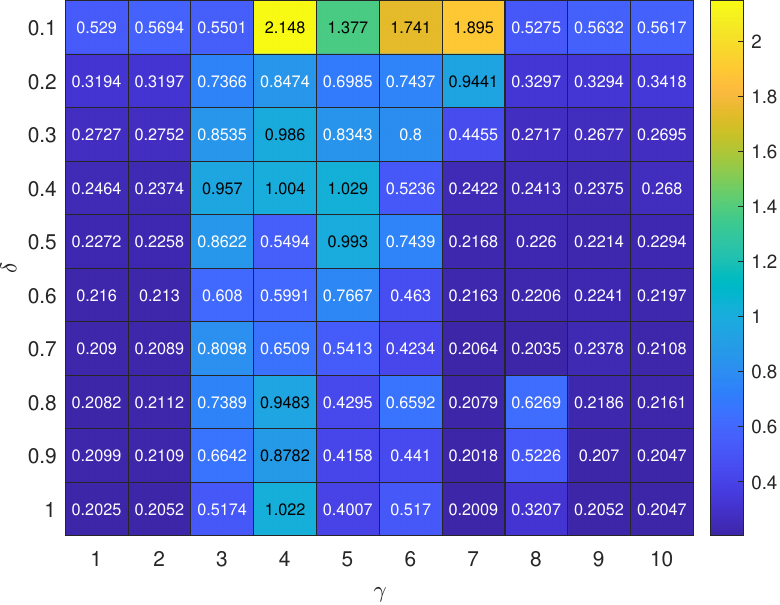}\label{Fig_parameter_analysis_TC_D10}}
  \caption{Parameter analysis of the proposed method on four selected datasets. 
  (a)-(d) The clustering accuracy (\%) of GBDPC on datasets D7-D10.
  (e)-(h) The computing overhead (in seconds) of GBDPC on datasets D7-D10.}
  \label{Fig_Parameter_Sensitivity_Analysis}
\end{figure*}

We proceed to evaluate the performance of our proposed method on twelve publicly available datasets, the details of which are provided in Table \ref{Table_real_datasets}. These datasets were randomly selected from the UCI Machine Learning Repository \footnote{http://archive.ics.uci.edu/ml}.
We start by comparing our GB generation method with those proposed by \cite{ChengDongDong2023} and \cite{XieJiang2023} through two GB-based clustering algorithms: GBDPC and GBSC. The GB generation method in the GBDPC (resp. GBSC) algorithm is replaced by the GB generation methods proposed in this paper and \cite{XieJiang2023} (resp. \cite{ChengDongDong2023}). Together with existing GBDPC and GBSC algorithms, we obtain six distinct versions of GB-based clustering algorithms. These algorithms are then applied to the twelve publicly available datasets. Additionally, we compare them with non-GB-based clustering algorithms, including DPC \cite{Rodriguez2014}, SC \cite{VonLuxburg2007395}, k-means++ \cite{Arthur20071027}, ultra-scalable spectral clustering (U-SPEC) \cite{HuangDong2020}, and ultra-scalable ensemble clustering (U-SENC) \cite{HuangDong2020}. The parameter settings for these algorithms are detailed in Table \ref{Table_parameters_real_dataset}, where $\sigma$ represents the parameter used to calculate instance similarity in SC and GBSC, and $\lambda$ is the parameter used to determine the truncation distance of DP\footnote{Let $md$ be the maximum distance between instances. Then the truncation distance of DP is given by $\lambda\cdot md$.}. Additionally, $\gamma$ and $\delta$ are parameters specific to our proposed method. All of these parameters are optimized using a grid search method to maximize clustering accuracy. The clustering results are evaluated using three indicators: clustering accuracy \cite{YanDonghui2009}, normalized mutual information (NMI) \cite{Slonim2000617}, and computing overhead.

The experimental results, presented in Table \ref{Table_ACC_TC_Real_Datasets} and Fig. \ref{Fig_RealDatasetACCNMI}, reveal several key findings. Firstly, utilizing our proposed method to generate GBs leads to GBDPC (resp. GBSC) achieving the highest clustering accuracy on six (resp. three) publicly available datasets. Secondly, similar results are obtained from the NMI perspective. Specifically, GBDPC (resp. GBSC) achieves the highest NMI on five (resp. four) publicly available datasets by employing our method. Notably, the GBSC, employing our GB generation method, achieves the highest average clustering accuracy and NMI across all twelve publicly available datasets. Thus, compared to the other two GB generation methods, our approach yields superior clustering accuracy and NMI. Thirdly, among the three GB generation methods, the average computing overhead of GB-based clustering algorithms generated by Cheng et al.'s (resp. Xie et al.'s) method is the lowest (resp. highest). Therefore, employing our method for GB generation not only yields the highest average clustering accuracy and NMI but also offers advantages in terms of computing overhead.  It is noteworthy that in Table \ref{Table_parameters_real_dataset}, the parameter $\delta$ is equal to 1 in only three out of twenty-four cases. Hence, to maximize clustering accuracy, it is essential to incorporate the parameter $\delta$ into our proposed method.

\subsection{Parameter Analysis}

In the experiment results presented in subsection \ref{Section_Real_Datasets}, different parameter combinations were employed in our proposed method to maximize clustering accuracy (see Table \ref{Table_parameters_real_dataset}). Therefore, it is crucial to further analyze the impact of different parameter combinations on the proposed method. In this subsection, we focus on publicly available datasets D7-D10 and the GBDPC algorithm to illustrate the impact of various parameter combinations on clustering accuracy and computing overhead. The experiment results are depicted in Fig. \ref{Fig_Parameter_Sensitivity_Analysis}.

Firstly, it is evident that changes of the parameter $\gamma$ have little impact on clustering accuracy. Particularly noteworthy, for datasets D7 and D8, the variation of $\gamma$ has no discernible effect on clustering accuracy (see Figs. \ref{Fig_parameter_analysis_ACC_D7} and \ref{Fig_parameter_analysis_ACC_D8}). However, the selection of the parameter $\delta$ plays a critical role in clustering accuracy. Optimal selection of $\delta$ leads to relatively high clustering accuracy, underscoring the importance of incorporating the parameter $\delta$.

Secondly, as depicted in Figs. \ref{Fig_parameter_analysis_TC_D7}-\ref{Fig_parameter_analysis_TC_D10}, the computing overhead tends to decrease with an increase in $\delta$. Notably, among 100 parameter combinations, the maximum computing overhead consistently occurs when $\delta$ is set to $0.1$. Moreover, for the datasets D9 and D10, the computing overhead exhibits a trend of initially increasing and then decreasing with the rise of the parameter $\gamma$. However, such a phenomenon is not observed in the datasets D7 and D8.

Consequently, based on the above analysis, it can be concluded that selecting the parameter $\delta$ is pivotal for clustering accuracy. Additionally, although the parameter $\gamma$ minimally affects clustering accuracy, its selection may significantly influence the computing overhead.

\section{Conclusion}\label{Section5}
In this article, we propose a novel and 
effective GB generation method for clustering tasks based on the POJG. Compared 
with some existing methods, the proposed method has the following 
advantages: 
\begin{enumerate}
  \item [(1)] Our method utilizes the POJG to comprehensively evaluate the quality of GBs and detect abnormal ones, which guarantees the rationality of the generated GBs.
  \item [(2)] By employing a binary tree pruning-based strategy to compute the best combination of sub-GBs, our method generates GBs that better align with the data distribution compared to threshold-based or greedy strategies.
  \item [(3)] The GB-based clustering algorithms employing our GB generation method achieve superior clustering accuracy and NMI.
  to generate GBs.
\end{enumerate}

In the future, it would be intriguing to explore methods for efficiently and effectively determining the optimal parameter combinations for our proposed approach. Additionally, we aim to further investigate the integration of our proposed GB generation methods with other unsupervised learning techniques, which will be helpful for discovering meaningful patterns in data. Moreover, it may be worthwhile to consider constructing GBs using alternative distance metrics besides Euclidean distance, which will imply different geometries of information granules.

\ifCLASSOPTIONcaptionsoff
  \newpage
\fi



\bibliography{reference}
\bibliographystyle{IEEEtran}

\end{document}